\documentclass[10pt,twocolumn,letterpaper]{article}

\usepackage{iccv}              
\usepackage{times}
\usepackage{epsfig}
\usepackage{graphicx}
\usepackage{amsmath}
\usepackage{amssymb}
\usepackage{microtype}
\usepackage{inconsolata}
\usepackage{makecell} 
\usepackage{colortbl} 
\usepackage{stfloats}
\usepackage{float}
\usepackage{xcolor}
\usepackage{multirow}
\usepackage{subcaption}
\usepackage{xcolor}  
\usepackage{multirow}
\usepackage{booktabs} 
\usepackage{array}   
\usepackage{algorithm}
\usepackage{algpseudocode}
\usepackage[dvipsnames]{xcolor}
\usepackage{amssymb}
\usepackage{enumitem}
\usepackage{arydshln}
\usepackage{pifont}
\newcommand{\DialNav}{DialNav}

\newcommand{\Nav}{Navigator}
\newcommand{\Gui}{Guide}
\newcommand{\Task}{DialNav}
\newcommand{\Dataset}{RAIN}
\newcommand{\InstanceData}{RAIN-Segment}
\newcommand\Tstrut{\rule{0pt}{2.6ex}}       
\newcommand\Bstrut{\rule[-1.2ex]{0pt}{0pt}} 
\newcommand{\cmark}{\ding{51}}%
\newcommand{\xmark}{\ding{55}}%

\definecolor{iccvblue}{rgb}{0.21,0.49,0.74}
\usepackage[pagebackref,breaklinks,colorlinks,allcolors=iccvblue]{hyperref}
\title{DialNav: Multi-turn Dialog Navigation with a Remote Guide}

\author{
Leekyeung Han$^{1}$ \quad
Hyunji Min$^{1}$ \quad
Gyeom Hwangbo$^2$ \\
Jonghyun Choi$^{3}$ \quad
Paul Hongsuck Seo$^{1}$ \\
$^1$Korea University \quad
$^2$University of Seoul \quad
$^3$Seoul National University\\
{\tt\small \{happilee12, daream2, phseo\}@korea.ac.kr, gyeomhwangbo@uos.ac.kr, jonghyunchoi@snu.ac.kr}
}

\begin{document}

\twocolumn[{
\renewcommand\twocolumn[1][]{#1}
\maketitle
\begin{center}
\includegraphics[clip, width=1.0\linewidth]{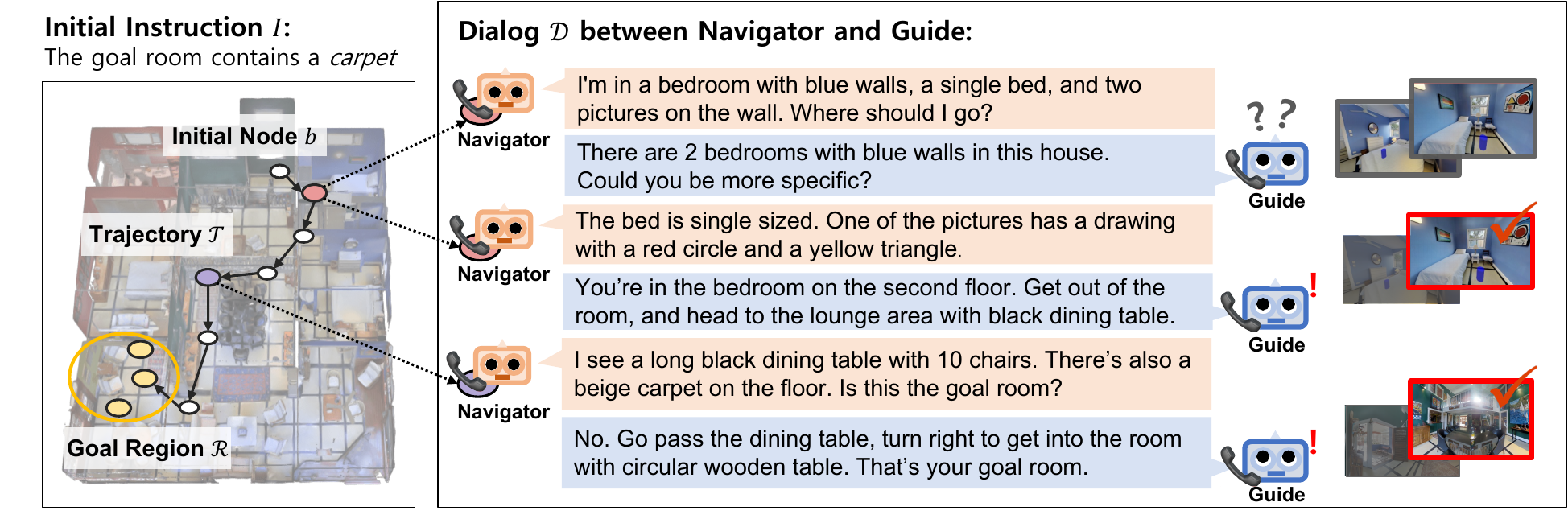}
\captionof{figure}{\textbf{Overview of the proposed \Task{} task.}
\Nav{} is tasked with reaching the goal region $\mathcal{R}$ (yellow circle in the map) from the initial node $b$ based on an ambiguous initial instruction $I$, which provides only a hint about $\mathcal{R}$ (\textit{e.g.}, `\textit{The goal room contains a carpet}').
During navigation, \Nav{} can ask questions to obtain additional guidance (orange text boxes). 
\Gui{} has knowledge of the environment but lacks information of \Nav{}'s location.
For successful navigation, \Gui{} must estimate \Nav{}'s location through dialog $\mathcal{D}$ and provide directions from the estimated position to the goal region (blue text boxes).
Note that each QA pair in $\mathcal{D}$ is mapped to a node (red and purple nodes mapped to dialog turns by dotted arrows) in the navigation trajectory $\mathcal{T}$.
\label{fig:main-figure}
}
\end{center}
}]

\let\thefootnote\relax

\begin{abstract}
We introduce DialNav, a novel collaborative embodied dialog task, where a navigation agent (\Nav{}) and a remote guide (\Gui{}) engage in multi-turn dialog to reach a goal location.
Unlike prior work, DialNav aims for holistic evaluation and requires the \Gui{} to infer the \Nav{}'s location, making communication essential for task success.
To support this task, we collect and release Remote Assistance in Navigation (RAIN) dataset, human-human dialog paired with navigation trajectories in photorealistic environments. 
We design a comprehensive benchmark to evaluate both navigation and dialog, and conduct extensive experiments analyzing  the impact of different Navigator and Guide models. 
We highlight key challenges and publicly release the dataset, code, and evaluation framework to foster future research in embodied dialog.
Our code and dataset are available at: \href{https://happilee12.github.io/DialNav}{https://happilee12.github.io/DialNav}.
\end{abstract}

\section{Introduction}

An embodied AI agent with a physical body perceives and interacts with the world, responding to human instructions or environmental stimuli. 
Since an embodied agent physically engages with its surroundings, misinterpreting human commands or hallucinating unintended actions can cause inconvenience or even physical harm. 
To mitigate these risks, the agent must seek clarification when faced with ambiguous tasks.
A natural approach is dialog-based interaction, allowing the agent to refine its understanding before acting. 
This enhances both safety and effectiveness in task execution.

Despite the importance of communicative ability in embodied AI, its inherent challenges hinders progress.
First, collecting dialog-based data is costly, as it requires two individuals to engage in real-time interaction while performing the task. 
Secondly, a dialog-based task necessitates not only equipping the agent with the ability to ask questions but also constructing a counterpart model capable of providing responses. 
Additionally, a framework that enables task execution through question-and-answer interactions is required. 
Lastly, the dynamic and interdependent nature of the task makes it difficult to evaluate performance. 
Due to these challenges, prior research has predominantly focused on task execution rather than leveraging dialog for task completion.

Some studies have already acknowledged the importance of dialog in embodied AI and has explored tasks involving dynamic additional instructions from a guide~\cite{nguyen2019help, nguyen2019vision, thomason2020vision, gao2022dialfred, padmakumar2022teach}. 
However, these studies often assume that the guide has full knowledge of current situation—an unrealistic assumption in practical settings. 
This omniscient guide model reduces the incentive for the guide to carefully consider the performer’s questions and, in turn, discourages the performer from formulating high-quality inquiries.
Even a vague request like ``help" can elicit a perfect response from an all-knowing guide, making the dialog less meaningful.

To address this limitation, we propose \Task{}.
\Task{} is a cooperative navigation task between \Nav{} and Remote \Gui{}, where \Gui{} is not aware of \Nav{}'s location but only familiar with the environment.
This setting closely mirrors real-world scenarios, such as when a lost person calls a friend for directions. 
Here, \Nav{} must formulate high-quality, detailed questions, as the effectiveness of \Gui's guidance depends on the clarity of the queries. 
Likewise, \Gui{} is incentivized to pay close attention to \Nav’s questions, as they lack direct awareness of \Nav's location.
Some prior works~\cite{banerjee2021robotslang, devries2018talkwalknavigatingnew} have explored non-omniscient guides, but under restricted conditions—\cite{banerjee2021robotslang} evaluates agents using static dialog histories in a physical maze, while \cite{devries2018talkwalknavigatingnew} limits the guide's view to a 2D grid map, reducing spatial fidelity.
In contrast, \Task{} supports long-horizon dialog and navigation in realistic indoor environments, enabling richer and more purposeful communication.

To support this task, we introduce the Remote Assistance in Navigation dataset (\Dataset), a visual navigation dataset with multi-turn human-human dialog.
To this end, we developed a data collection tool, specifically designed to capture complex interactions in \Task{}. 
As \Dataset{} dataset can be used to train and evaluate the core capabilities of \Nav{} and \Gui{} under standardized conditions, we conduct experiments with \Nav{} and \Gui{} models across various setups and module configurations, providing valuable insights into this new task.
We will publicly release the complete codebase, data collection tool, dataset, and training and evaluation frameworks, ensuring reproducibility and providing foundational resources for future research.
Our main contributions are summarized as follows:

\begin{itemize} 
    \item We introduce a cooperative embodied dialog navigation task involving \Nav{} and remote \Gui{}.    
    \item We collect and release \Dataset{}, which captures human-human interactions for \Task{}.
   \item \Nav{} and \Gui{} are trained and tested under various setups within a holistic dialog-and-navigation pipeline.
    \item We discuss key challenges and their implications for fostering future research.
\end{itemize}

\section{Related Work}
\noindent \textbf{Embodied Dialog} \ \ 
Recent advancements in embodied dialog have led to the development of new benchmarks and datasets across question answering~\cite{das2018embodied, majumdar2024openeqa}, navigation~\cite{thomason2020vision, fan-etal-2023-aerial}, and manipulation~\cite{gao2022dialfred, padmakumar2022teach, suhr2019, narayan-chen2019}.
Although these benchmarks incorporate dialog, they either assume an omniscient guide~\cite{roman2020rmm, nguyen2019help, gao2022dialfred, padmakumar2022teach, narayan-chen2019} or restrict evaluation to task execution only~\cite{thomason2020vision, banerjee2021robotslang, hahn2020you}, thereby underestimating the role of dialog (\textit{e.g.}, asking for help without a question~\cite{nguyen2019help}).
Our work models two dialog-enabled agents, \Nav{} and remote \Gui{}, and emphasizes the importance of dialog generation and understanding within the full communication-action loop.
The most closely related work is Talk the Walk~\cite{devries2018talkwalknavigatingnew} but it operates in a highly constrained setting with limited agent actions and simplified 2D grid map for the guide.
In contrast, our task supports long-horizon, multi-turn dialog in a realistic navigation environments.

\noindent \textbf{Vision and Language Navigation} \ \ 
Vision-and-Language Navigation (VLN) is a prominent multimodal embodied task where an agent navigates a visual environment based on language instructions.
\citet{anderson2018vision} pioneered the field by introducing the first photo-realistic VLN task, dataset, and a simulator.
Since then, a wide range of VLN benchmarks have emerged, but they often face limitations, either providing overly detailed instructions~\cite{jain2019stay, ku2020room} unnatural for human interaction or offering ambiguous guidance~\cite{qi2020reverie,zhu2021soon} insufficient for successful navigation.
To overcome these limitations, this study explores a realistic dialog-based navigation task.

\noindent \textbf{Subtasks for \Task{}} \ \
\Task{} targets the holistic evaluation of the communication-and-navigation process, integrating multiple subtasks that have previously been addressed only in isolation.
Instruction generation~\cite{fried2018speaker, zhu2020multimodal, yan2024instrugenautomaticinstructiongeneration, wang2022counterfactual, wang2023lana, wang2022less, Chen_2022_HM3D_AutoVLN, fan2023r2h}, explored primarily for data augmentation in VLN, is closely related to the question and answer generation components of \Task{}.
Navigator localization is a distinguishing aspect of \Task{} compared to previous VDN tasks. 
Unlike SLAM~\cite{davison2003real} focusing on self-localization, this task identifies the agent location based on natural language description.
The Localization from Embodied Dialog (LED) task was introduced in~\cite{hahn2020you}, followed by a few subsequent studies~\cite{zhang2024dialoc, hahn2022transformer, pate2023wallpaper, li2024topviewrs, wang2025towards}.
While navigator localization in \Task{} shares similarities with LED, it presents additional challenges due to its dynamic navigation context and long-form dialog.
Deciding when to ask for help has been explored in prior work~\cite{singh2022ask4help, zhu2021self, gao2022dialfred}, often by introducing an auxiliary model or leveraging action space entropy.
We adopt these models for the corresponding modules in \Task{}.

\begin{figure*}[t]
    \centering
    \begin{subfigure}{0.32\textwidth}
        \centering
        \includegraphics[width=\linewidth]{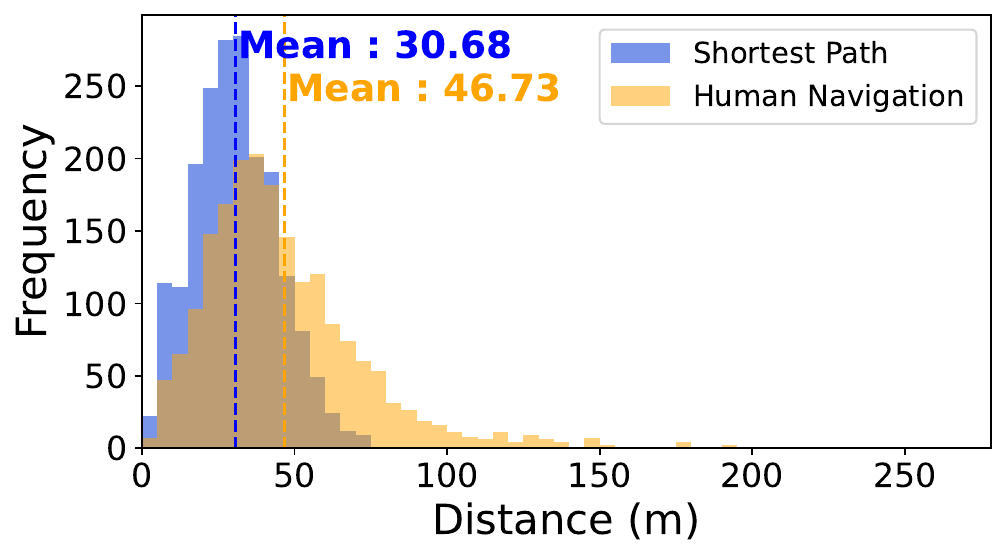}
    \end{subfigure}
    \hfill
    \begin{subfigure}{0.32\textwidth}
        \centering
        \includegraphics[width=\linewidth]{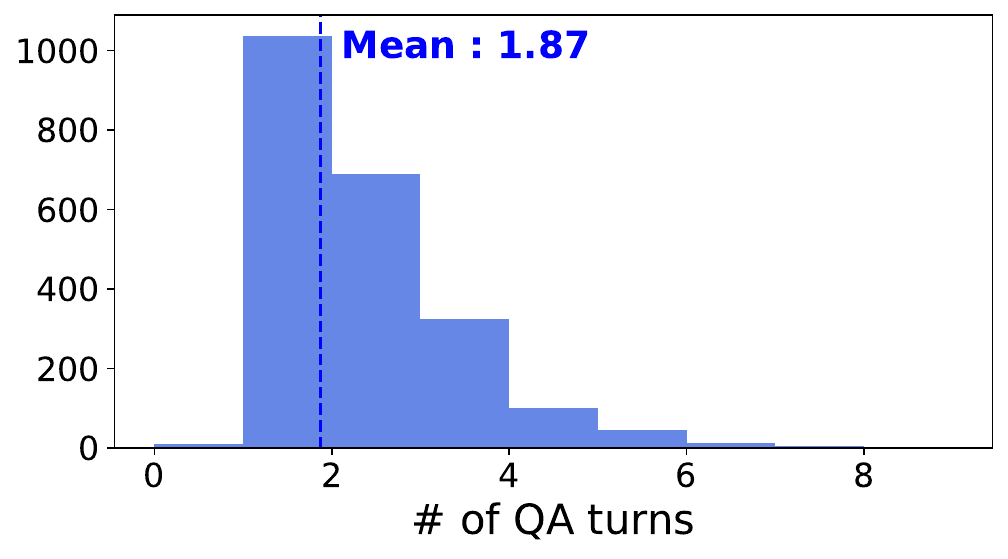}
    \end{subfigure}
    \hfill
    \begin{subfigure}{0.32\textwidth}
        \centering
        \includegraphics[width=\linewidth]{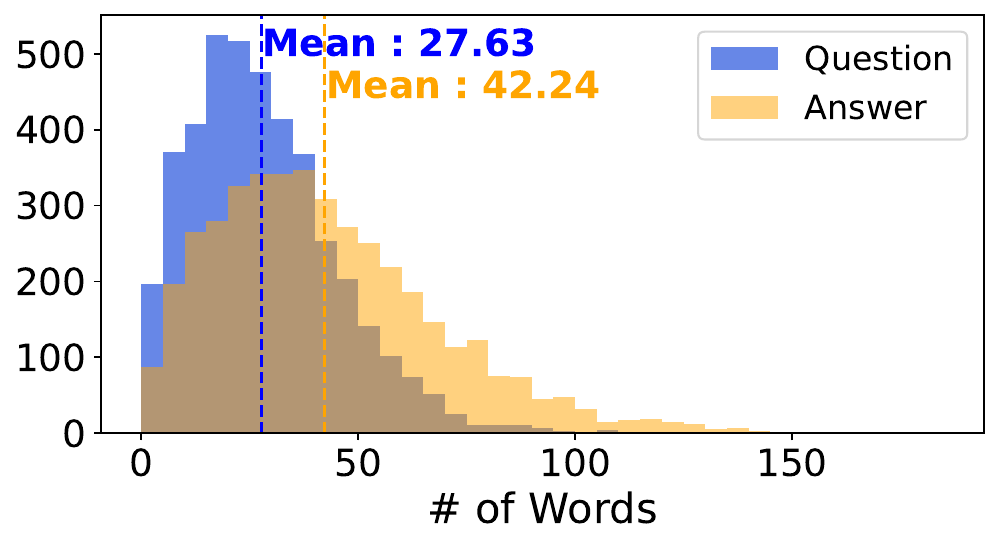}
    \end{subfigure}
    \vspace{-0.3cm}
    \caption{\textbf{Data distributions in RAIN.} Distributions of trajectory lengths for shortest paths and human navigation (left), QA pairs per episode (center), and word counts per question and answer (right).
    }
    \vspace{-0.4cm}
    \label{fig:statistics}
\end{figure*}

\section{Task and Dataset}

\subsection{The \Task{} Task} 
\Task{} is a novel navigation task that targets the holistic evaluation of the dialog and execution loop, where an agent (\textit{\Nav{}}) navigates to a specified goal region with assistance from a remote guide (\textit{\Gui{}}).
Unlike prior studies~\cite{nguyen2019help,nguyen2019vision,singh2022ask4help,roman2020rmm,gao2022dialfred,padmakumar2022teach}, \Task{} incorporates a remote guide and holistic evaluation, encompassing all key subtasks for dialog-enabled navigation. 

In \Task{}, \Nav{} starts with an ambiguous instruction and gathers additional information through natural language dialog with \Gui{}.
\Nav{} must decide when to seek assistance and generate relevant questions.
\Gui{}, upon receiving a query, infers \Nav{}'s location before providing a response, incentivizing \Nav{} to ask informative questions.
Dialog is initiated by \Nav{} and follows an alternating turn-taking format, with no limit on QA turns.
\cref{tab:dataset_comparison} compares \Task{} to existing VLN tasks in photo-realistic environments.

\begin{table}[t]
    \centering
    \scalebox{0.85}{
        \begin{tabular}{lccccccc}
           \toprule
            &\multicolumn{5}{c}{\textbf{Subtasks}} \\
            \textbf{Tasks} &\textit{N} & \textit{W}  & \textit{Q} & \textit{L} & \textit{A} & \textbf{Guide} \\
            \cmidrule(r){1-1}
            \cmidrule(lr){2-4} 
            \cmidrule(lr){5-6} 
            \cmidrule(lr){7-7} 
            VLN~\cite{anderson2018vision} & \cmark & & & & & \xmark \\
            Just Ask~\cite{chi2019justaskaninteractivelearning} & \cmark & \cmark & & & & Omniscient \\
            HANNA~\cite{nguyen2019help} & \cmark & \cmark & & & & Omniscient \\
            VDN~\cite{thomason2020vision} & \cmark & \cmark & \cmark  & & \cmark & Omniscient \\
            \textbf{\Task (Ours)} & \cmark & \cmark & \cmark & \cmark & \cmark & Remote\\
            \bottomrule
        \end{tabular}
    }
    \vspace{-0.2cm}

\caption{\textbf{Comparing vision-and-language navigation tasks in photo-realistic environments.} The `Subtasks' column indicates tasks involved in \Task{}: 
\textit{N}: navigation, \textit{W}: deciding whether to ask, \textit{Q}: question generation, \textit{L}:  localizating navigator, \textit{A}: answer generation. Navigation, deciding whether to ask, and question generation are subtasks for \Nav{}, while localizing navigator and answer generation are subtasks for \Gui{}.
}
\vspace{-0.4cm}
\label{tab:dataset_comparison}
\end{table}

\subsection{Dataset Collection} 
To address \Task{}, we collected \Dataset{}, a dataset consisting of episodes in which two human annotators interact to complete a navigation task. 
This section outlines the dataset curation process and presents a statistical analysis of its key attributes.

\noindent \textbf{Navigation Simulator} \ \ 
Our study is based on the Matterport3D simulator~\cite{anderson2018vision}, which utilizes photo-realistic reconstructions of real-world houses.
The simulator represents each reconstructed house as a graph $G = (\mathcal{V}, \mathcal{E})$, where $\mathcal{V}$ and $\mathcal{E}$ denote a set of navigable nodes representing locations within the house and a set of edges connecting nodes, respectively.
An agent can traverse over the environment graph $G$ moving between nodes $v\in \mathcal{V}$ connected by edges $e\in \mathcal{E}$.
The agent is given a panoramic view at each node.
We use 83 house scans partitioned into 61 for train and val seen, 11 for val unseen and 18 for test respectively following~\cite{anderson2018vision}.
The distinction between val seen and val unseen is based on whether the house is included in the training split.

\noindent \textbf{Episodes} \ \ 
RAIN contains a total of 2,231 DialNav episodes. 
Each episode $E=(G, b, \mathcal{R}, I, \mathcal{T}, \mathcal{D})$ is a 6-tuple composed of an environment graph $G$, \Nav{}'s initial node $b\in \mathcal{V}$, a set of nodes $\mathcal{R} \subset \mathcal{V}$ in the goal region, an initial instruction $I$, \Nav{}'s navigation trajectory $\mathcal{T}$ from $b$ to one of goal nodes $r\in\mathcal{R}$, and the dialog $\mathcal{D}$ between \Nav{} and \Gui{}.
Note that the first four elements $(G, b, \mathcal{R}, I)$ define a navigation task and a goal region (\textit{e.g.} \textit{living room} or \textit{bathroom}) may contain multiple nodes ($|\mathcal{R}|\ge 1$).
Our dataset includes 1,401 navigation tasks from CVDN~\cite{thomason2020vision} and 838 additional tasks generated following the same scheme.
The trajectory $\mathcal{T}$ and dialog $\mathcal{D}$ are collected from human annotators.
A trajectory $\mathcal{T}=(v_1, v_2, ..., v_N)$ is a sequence of nodes $v_i \in \mathcal{V}$ with $v_1 = b$ and $v_N \in \mathcal{R}$.
A dialog $\mathcal{D}=(d_1, d_2,...,d_M)$ where $d_i=(q_i, a_i, u_i)$ is a triplet of a question $q_i$ posed by \Nav{}, its corresponding answer $a_i$ from \Gui{} and the node $u_i$ where this question-answer exchange occurs.
The node $u_i$  represents the last visited node in the trajectory at the corresponding dialog turn.

\noindent \textbf{Data Collection Tool} \ \ 
We developed a data collection tool to simulate \Task{}. 
It provides \Nav{} with a navigation interface and a real-time chat interface for communication with \Gui{}.
Meanwhile, unlike previous studies that provide the guide with next optimal steps from \Nav{}'s exact location, our tool requires \Gui{} to infer \Nav{}'s position using a specialized interface that displays the entire house layout and room list.
Once \Gui{} selects a node, the tool provides the shortest path trajectory to the goal from that node, assisting in answer generation.
For each episode, two annotators are randomly assigned as \Nav{} and \Gui{}, and given a task $(G, b, \mathcal{R}, I)$.
More details about our tool can be found in \cref{sec:data_collection_interface} in Supp. Mat.
We will release the code for the collection tool to support future research.

\begin{table*}[t]
    \centering
    \scalebox{0.85}{
    \begin{tabular}{p{0.20\textwidth}>{\centering\arraybackslash}p{0.05\textwidth}>{\centering\arraybackslash}p{0.05\textwidth}p{0.75\textwidth}}
    \toprule
    \textbf{Dialog characteristics}  & \textbf{Init.} & \textbf{Subs.} & \textbf{Example} \\
    \midrule
    \textit{Nav} describes scene & 0.97 & 0.74 & 
    \parbox[l][2.5em][l]{0.74\textwidth}{\textit{Nav}: \textbf{I'm the living room which is connected to the kitchen area. There's two beige arm chairs and one beige soft with beige and blue colored cushions on it. ...}}\\
    \addlinespace[0.5em]
    \textit{Gui} describes path & 0.86 & 0.75 &
    \parbox[c][2.5em][c]{0.74\textwidth}{\textit{Gui}: \textbf{go upstairs; turn right once you're upstairs and go straight along the hallway. To your left, there should be an open door with a circular table on top of a large carpet.}  ...} \\ 
    \addlinespace[0.5em]
    \textit{Gui}  requests clarification & 0.15 & 0.03 &
    \parbox[c][1.5em][c]{0.74\textwidth}{\textit{Gui}: By any chance, \textbf{do you see a swimming pool?}} \\
    \addlinespace[0.5em]
    \textit{Nav}  requests clarification & - & 0.13 &
    \parbox[c][1.5em][c]{0.79\textwidth}{\textit{Nav}:  sharp left turn,  \textbf{you mean go through between the stove and wooden big table ?}}\\
    \addlinespace[0.5em]
    Confirms goal & - & 0.14 &
    \parbox[c][2.5em][c]{0.79\textwidth}{\textit{Nav}: ... there is a statue of woman in front of the long wooden table. \textbf{Am I at the goal room?} \\ \textit{Gui}: \textbf{Yes correct.} good work}\\
    \addlinespace[0.5em]
    Need dialog history & - & 0.46  &
    \parbox[c][1.5em][c]{0.79\textwidth}{\textit{Nav}: yea now i'm standing in front of \textbf{those two statues}}\\
    \addlinespace[0.3em]
    \bottomrule
    \end{tabular}
    }
    \captionsetup{font=small}
    \vspace{-0.2cm}
    \caption{{
        \textbf{Dialog characteristics in \Dataset{}.} 
        We manually analyzed 100 randomly sampled episodes from RAIN and present various dialog characteristics in \Dataset{}, along with an example.
        The Init. and Subs. columns indicate the frequencies of these characteristics in the first and subsequent dialog turns, respectively.
    }
    \label{table:additional_features_of_qa}
    }
    \vspace{-0.3cm}
   
\end{table*}
\begin{figure}[t]
    \centering
    \footnotesize

    \begin{minipage}{1\linewidth}
        \centering
        \begin{tabular}{|cp{0.8\linewidth}|}
            \hline
            \Tstrut Q: & Think I'm at the room what you've noticed me. On my left side, there is a table made of transparent glass and on my right side, there is a Queen side bed. Also with the two doors opened to the balcony. \\
            \Tstrut A: & That's right. Go out the left door to the balcony, you'll see two black stools. That balcony is your target room.  \\
            \hline
        \end{tabular}
        \subcaption{\Dataset{}}
    \end{minipage}%
    \vspace{0.2cm}
    \begin{minipage}{1\linewidth}
        \centering
        \begin{tabular}{|cp{0.8\linewidth}|}
            \hline
            \Tstrut Q: & Should I exit this room? \\ 
            \Tstrut A: & Yes, go out blue door. \Bstrut \\
            \hline
        \end{tabular}
        \subcaption{CVDN}
    \end{minipage}
    \vspace{-0.3cm}
    \caption{\textbf{Example QA pairs from \Dataset{} and CVDN.}
    RAIN questions include detailed environmental descriptions to help the remote guide infer \Nav{}'s location.
    In contrast, CVDN questions simply request the next action, as detailed descriptions are unnecessary for an omniscient guide.
    Additional examples from RAIN are provided in Supp. Mat.
    }
    \label{fig:rain-qa-sample}
    \vspace{-0.3cm}
\end{figure}

\noindent \textbf{Data Collection Process} \ \ 
To ensure smooth data collection, annotators first watched a tutorial video and completed two practice episodes beforehand.
All annotators provided informed consent for data usage in research.
To ensure high-quality dataset, human annotators evaluated each other on a 5-point scale after completing each episode.
The average scores are 4.48 and 4.28 for Navigator and Guide respectively. 
We imposed a time limit of 22 minutes for each episode, ensuring that only efficient plays were included. One episode took 8 minutes in average.
The final dataset comprises 2,231 episodes, distributed across training (1,559), validation seen (111), validation unseen (276), and test (285) sets.
The total cost for data collection was approximately 7,500 USD.
Refer to \cref{sec:data_collection_details} in Supp. Mat. for further details.

\subsection{Statistics}

\noindent \textbf{Trajectory} \ \ 
\cref{fig:statistics}(left) compares trajectory length distributions between shortest paths and human navigation in \Dataset{}. 
Shortest paths range from 2.87m to 76.55m (avg. 30.68m with 17.39 nodes), while human navigators travel 3.02m to 262.64m (avg. 46.73m with 25.97 nodes), making human trajectories 1.62 times longer on average. 
The distribution is long-tailed, with over 80\% deviating less than twice the shortest path length.
The largest deviation is 33.5 times longer than the shortest path. 
Large deviations in general are attributed to \Nav{}'s excessive exploration, especially when the ground-truth trajectory is short.

\noindent \textbf{Dialog} \ \ 
\cref{fig:statistics}(center) presents the distribution of QA turns in \Dataset{}. 
Each episode contains an average of 1.87 QA pairs, with over 92\% concluding within three QA pairs.
Ten episodes include no dialog, while the longest interaction consists of eight QA pairs.
As shown in \cref{fig:statistics}(right), questions and answers in \Dataset{} average 27.63 and 42.24 words, respectively. 
Since \Gui{} has extensive knowledge of the environment and each QA turn requires localizing \Nav{}, which is a costly process, \Gui{} tends to provide comprehensive responses when possible, leading to fewer QA turns but longer answers.

\noindent \textbf{\Dataset{} Dialog Features} \ \ 
\cref{table:additional_features_of_qa} highlights key dialog features in \Dataset{}.
\Task{} demands detailed dialog and continuous verification to ensure \Nav{} and \Gui{} stay aligned.
\Nav{} provides rich descriptions of its surroundings when asking questions (Row 1), while \Gui{} sometimes seeks clarification for unclear queries (Row 3), emphasizing the importance of precise questions in \Task{}.
This contrasts with CVDN~\cite{thomason2020vision}, where \Gui{} gives directions based solely on perfect knowledge of \Nav{}'s location (\cref{fig:rain-qa-sample}). 
Additionally, due to the ambiguous initial instruction in \Task{}, \Nav{} and \Gui{} engage in extra dialog upon reaching the goal for confirmation (Row 5).

\section{\Nav{} and \Gui{} Models}
\begin{figure}[t]
    \centering
     \includegraphics[width=0.95\linewidth]{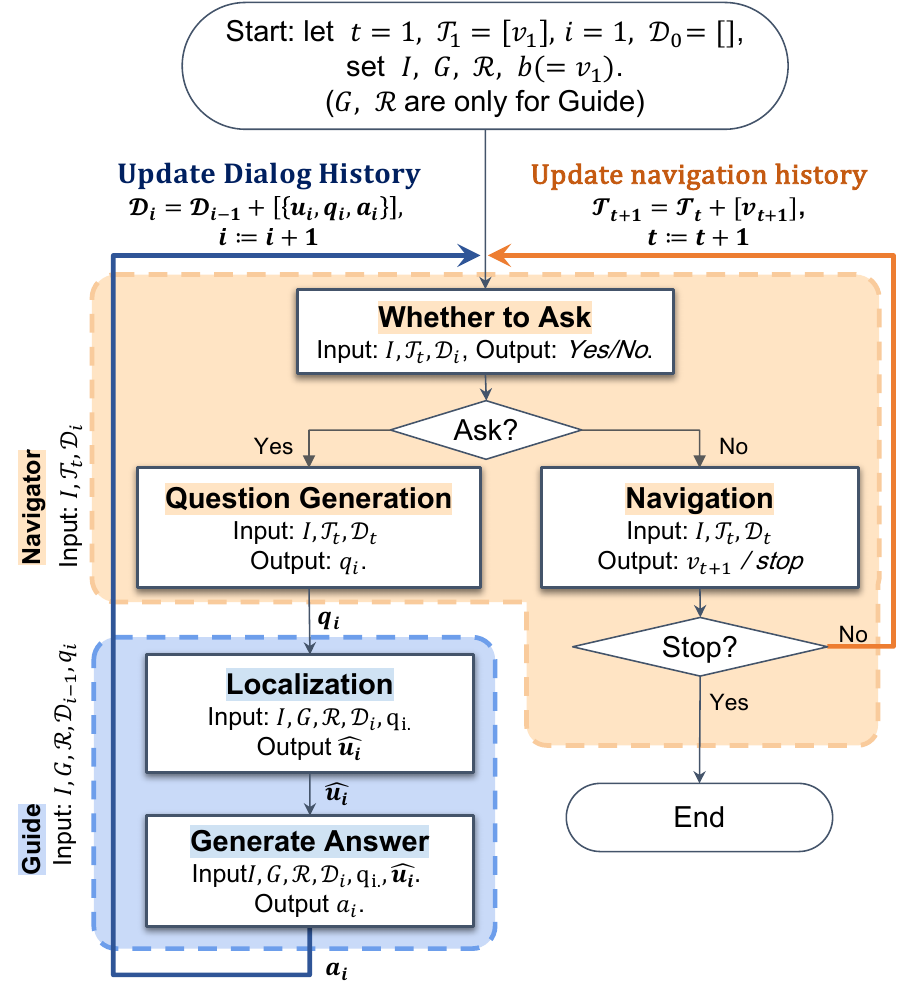}
    \vspace{-0.3cm}
    \caption{\textbf{Sequential interaction between \Nav{} and \Gui{}.} This diagram illustrates the collaborative execution of \Task{}. At each iteration, \Nav{} determines whether to ask a question. If a question is posed, \Gui{} first localizes \Nav{} and then formulates a response. Each question-answer exchange updates the dialog history. If no question is asked, \Nav{} proceeds with navigation. The task concludes when \Nav{} decides to stop.
    $\hat{u_{i+1}}$ represents the viewpoint where \Gui{} estimates \Nav{} to be based on localization.
    }
    \vspace{-0.4cm}
    \label{fig:task-diagram}
\end{figure}

The collaborative nature of \Task{} involves two models: \Nav{} and \Gui{}.
\Nav{} is a dialog-enabled agent that engages in conversation to gather additional navigation guidance, while \Gui{} is a remote assistant with knowledge of the environment, responding to \Nav{}'s inquiries.
This section introduces the core capabilities of \Nav{} and \Gui{}. 
\cref{fig:task-diagram} illustrates the overall process of \Task{}, highlighting the interaction between \Nav{} and \Gui{}.
To study these capabilities, we modularize each component and leverage existing models for different functionalities.

\subsection{Navigator} 
\label{sec:nav}

\Nav{} begins the task with an initial instruction $I$ and, at each timestep, either updates $\mathcal{T}_t$ by taking a navigation action or updates $\mathcal{D}_i$ by posing a question.
To effectively complete the task, \Nav{} requires modules, each handling one of the three core capabilities:
(1)~determining navigation actions,
(2)~identifying appropriate moments to ask questions, and (3)~generating relevant questions.

\noindent \textbf{Navigation} \ \ 
At each navigation step $t$, \Nav{} selects the next node based on past dialog and navigation history.
This is analogous to Vision-and-Language Navigation (VLN)~\cite{anderson2018vision}, where an agent navigates to a goal based on a natural language instruction.
Unlike VLN, where instructions are static and provided at the beginning, dialog in \Task{} is dynamically collected across multiple visited nodes.
We train a navigation module using a VLN model architecture, addressing this discrepancy by treating the past dialog as a single instruction for the remaining navigation path similarly to~\cite{thomason2020vision}.
Specifically, we evaluate HAMT~\cite{chen2021history} and DUET~\cite{chen2022think}, including pretrained weights from ScaleVLN~\cite{wang2023scaling}.
DUET with pretrained weights, which showed optimal performance, is adopted as our baselines.
Further details on model selection for each module are provided in \cref{sec:module-wise_training} in Supp. Mat.

\noindent \textbf{Whether to Ask} \ \ 
Asking questions at appropriate moments is another capability for \Nav{}; 
frequent questions overwhelm \Gui{}, while too few can lead to navigation errors.
Striking a balance between navigation performance and dialog efficiency is a crucial mission in dialog-based navigation.
In this work, we tested 3 approaches; Fixed-Interval~\cite{roman2020rmm}, where questions are asked at regular navigation step intervals, Confidence Thresholding~\cite{zhu2021self,gao2022dialfred}, where questions are triggered when action confidence falls below a threshold, and Decision Head~\cite{singh2022ask4help}, which learns questioning timing from \Dataset{} using action decision features.
The baseline model adopts Decision Head.

\noindent \textbf{Question Generation} \ \ 
When uncertain about the next navigation step, \Nav{} asks \Gui{} for additional information, often describing its surroundings to provide context.
For question generation, we use LANA~\cite{wang2023lana}, a VLN instruction generation model that, like this task, generates descriptions based on visual landmarks.
Additionally, we test LLaVA-1.5~\cite{liu2023llava}, a multimodal LLM, with prompts for comparison.
Although the target tasks differ, we pretrained LANA following~\cite{wang2023lana}, to generate navigation instructions.
LANA serves as our baseline module for question generation.

\subsection{Guide}
\Gui{} provides guidance to \Nav{} based on the environment graph $G$, initial instruction $I$, the goal region $\mathcal{R}$, and dialog history $\mathcal{D}_i$, which updates at each dialog turn $i$.
Note that since \Gui{} is remote, it does not have access to \Nav{}'s navigation history $\mathcal{T}_t$.

\noindent \textbf{Localization} \ \ 
When \Nav{} poses a question, \Gui{} must first estimate \Nav{}'s location.
This is similar to the Localization from Embodied Dialog (LED) task~\cite{hahn2020you}, which involves determining an observer's location based on dialog.
We evaluate two ranking models from~\cite{hahn2020you}: Simple Cross-Modal Network (SCN) and Graph Convolutional Network (GCN).
After testing both models pretrained on the WAY dataset~\cite{hahn2020you} and finetuned on \Dataset{}, we adopt GCN as our baseline module.

\noindent \textbf{Answer Generation} \ \ 
Answer generation involves formulating responses that guide \Nav{} toward the destination.
This primarily includes describing the next path to the goal region, similar to a VLN instruction generation.
We evaluate LANA~\cite{wang2023lana} and Llama-3.1-8B-instruct~\cite{touvron2023llama2openfoundation} with prompting for answer generation.
Like in question generation, we adopt LANA as our baseline module, initializing it with pretrained weights following~\cite{wang2023lana} and finetuning it on \Dataset{}.

\setlength{\tabcolsep}{5pt}
\begin{table}[t]
    \centering
\scalebox{0.85}{
    \begin{tabular}{lcccccccc}
    \toprule
    \textbf{Setup}
    & \textbf{SR$\uparrow$}
    & \textbf{OSR$\uparrow$}
    & \textbf{SPL$\uparrow$}
    & \textbf{NE$\downarrow$}
    & \textbf{NSC}
    & \textbf{DTC}
    & \textbf{LE$\downarrow$}
    \\
    \cmidrule(r){1-1}\cmidrule(lr){2-5}
    \cmidrule(lr){6-8}\cmidrule(lr){9-9}
    \multicolumn{9}{l}{\textit{Val Seen}} \\
    (1) Nav. only   &18.2 &28.6 &17.3 &14.5 &19.6 & -   & - \\
    (2) +Dialog     &27.0 &34.5 &25.4 &11.5 &16.4 &1.9 &22.6 \\
    (3) +GT Loc.    &\textbf{31.4} &\textbf{43.3} &\textbf{28.6} &\textbf{9.9} &18.3 &1.7 & - \\
    \midrule
    \multicolumn{9}{l}{\textit{Val Unseen}} \\
    (4) Nav. only   &15.4 &\textbf{33.3} &10.3 &14.9 &20.8 & -   & - \\
    (5) +Dialog     &13.9 &20.7 &10.0 &16.3 &14.2 &2.9 &21.9 \\
    (6) +GT Loc.    &\textbf{19.8} &29.0 &\textbf{15.5} &\textbf{12.7} &16.4 &2.6 & - \\
    \midrule
    \multicolumn{9}{l}{\textit{Test}} \\
    (7) Nav. only   &12.7 &\textbf{28.6} &8.0 &14.8 &22.8 & -   & - \\
    (8) +Dialog     &11.9 &19.0 &8.2 &17.3 &15.4 &2.9 &22.9 \\
    (9) +GT Loc.    &\textbf{17.3} &23.7 &\textbf{12.2} &\textbf{14.2} &17.4 &2.7 & - \\
    \bottomrule
    \end{tabular}
    }
    \vspace{-0.2cm}
    \caption{
    \textbf{Performance across dialog setup variations.} 
    In \textbf{Nav. only}, the agent~\cite{chen2022think} navigates solely based on the initial instruction $I$, mirroring the standard VLN setting and its performance in \Task{}.
    \textbf{+Dialog} enables dialog during navigation, corresponding to \Task{} setup.
    \textbf{+GT Loc.} provides \Gui{} with \Nav{}'s ground-truth location, removing localization error.}
    \label{tab:holistic-results}
    \vspace{-0.5cm}
\end{table}
\setlength{\tabcolsep}{6pt}

\section{Experiments}
\subsection{Experimental Settings}
\noindent \textbf{Training Modules} \ \ 
Every module in \Nav{} and \Gui{} is trained on \Dataset{}. 
During training, each episode is processed for its respective subtask (\textit{e.g.}, each QA turn forms a training example for question generation). 
We initialize each module with pretrained weights from related VLN tasks.
Further training details are provided in \cref{sec:module-wise_training} in Supp. Mat.
 
\noindent \textbf{Cooperative Evaluation} \ \ 
As depicted in \cref{fig:task-diagram}, \DialNav{} requires continuous collaboration between \Nav{} and \Gui{}.
Previously discussed modules are integrated in holistic dialog-based navigation.
We employ pretrained DUET~\cite{chen2022think} for navigation, LANA~\cite{wang2023lana} for question and answer generation and GCN for localization. All results reported with the average score of five runs. 

\noindent \textbf{Metrics} \ \ 
We report four navigation metrics adopted from prior VLN works~\cite{anderson2018vision, hahn2022transformer, wang2023scaling,hao2020towards}. 
\textbf{Success Rate (SR)} measures the percentage of episodes where \Nav{} successfully stops in the target region. 
\textbf{Oracle Success Rate (OSR)} counts episodes as successful if \Nav{} passes through the target region, benefiting from more exploration.
\textbf{Success weighted by Path Length (SPL)} evaluates the success rate while considering path efficiency, penalizing unnecessary detours.
\textbf{Navigation Error (NE)} measures the distance between \Nav{}'s final position and the goal, indicating how close \Nav{} ends up to the goal region.
Additionally, we report metrics for navigation and dialog efficiency, as well as localization accuracy.
\textbf{Navigation Step Count (NSC)} is the number of steps taken by the model, thereby measuring the extent of navigation exploration.
\textbf{Dialog Turn Count (DTC)} indicates the number of dialog turns in a single episode, reflecting the level of dialog engagement and efficiency.
\textbf{Localization Error (LE)}, measures the distance between \Gui{}'s predicted location and \Nav{}’s ground-truth position at each dialog turn, capturing cumulative errors from both question generation and localization.

\begin{table}
    \centering
    \scalebox{0.85}{
    \begin{tabular}{lccccc}
    \toprule
    \textbf{Non-pretrained module} 
    & \textbf{SR$\uparrow$}
    & \textbf{NE$\downarrow$}
    & \textbf{NSC}
    & \textbf{DTC}
    & \textbf{LE$\downarrow$}
    \\
    \cmidrule(r){1-1}\cmidrule(lr){2-3}
    \cmidrule(lr){4-6} 
    Fully pretrained & 27.0 & 11.4 & 16.4 & 1.9 & 22.6 \\
    $-$Navigation & 9.2 & 14.4 & 20.8 & 1.7 & 22.0 \\
    $-$Question generation & 27.7 & 11.0 & 16.4 & 1.9 & 20.8 \\
    $-$Localization & 25.7 & 12.8 & 16.8 & 2.0 & 25.1 \\
    $-$Answer generation & 11.7 & 14.8 & 16.9 & 1.9 & 23.5 \\
    \bottomrule
    \end{tabular}
    }
    \vspace{-0.2cm}
    \caption{
    \textbf{Effects of pretrained modules.} 
    Performance comparison when pretraining is omitted from specific modules.
    The `Non-pretrained' column indicates the module without pretrained initialization, while `Fully Pretrained' refers to the setup where all modules are pretrained.}
    \label{holistic-pretrain-variation}
    \vspace{-0.5cm}
\end{table}

\subsection{Results}

\noindent \textbf{Enabling Dialog} \ \ 
\cref{tab:holistic-results} compares models with and without dialog, highlighting its impact on visual navigation.
 In seen environments (Val Seen), the model without dialog (Row 1) relies solely on the ambiguous initial instruction, leading to poor navigation performance.
In contrast, incorporating dialog (Row 2) enables active engagement, as reflected in DTC (1.9 dialog turns per episode), facilitating disambiguation and informed decision-making. 
As a result, navigation performance improves significantly, with relative gains of 8.8\%, 5.9\%, 8.2\% and 3m in SR, OSR, SPL and NE respectively.
We further tested a setup where the GT location of \Nav{} is provided to \Gui{} when a question is posed. 
This simulates an omniscient guide, removing the need for well-formed questions and precise localization, similar to prior works~\cite{thomason2020vision}. 
The results show improved dialog efficiency with a lower DTC, as expected, and enhanced navigation performance due to more accurate answers enabled by perfect localization. 
These findings underscore the importance of well-posed questions and accurate localization in \Task{}.

\noindent \textbf{Generalization to Unseen Environments} \ \ 
The trends observed in unseen environments (Val Unseen and Test) differ from those in seen environments. 
Overall, performance is poor in both settings, with and without dialog.
We conjecture that this is due to the limited dataset scale, constrained by the high cost of data collection process compared to the complexity of \Task{}. 
Addressing this challenge remains a key direction for future research.
While both setups perform poorly, they exhibit different behaviors.
The model without dialog (Row 4 and 7) explores the environment excessively, resulting in an unnecessarily large number of navigation steps (NSC). 
This leads to high OSR (favoring exploration) but low SR and SPL, as the task inherently lacks sufficient information to determine the correct stopping location.

In contrast, the model with dialog (Row 5 and 8) continuously seeks additional information, as it remains uncertain about the unseen environment. 
This behavior significantly reduces unnecessary exploration. 
However, SR remains low when the question and answer generation modules fail to produce grounded dialog.
When the GT location of \Nav{} is given, navigation performance improves as the model no longer depends on  question generation and localization. 
By removing the reliance on these components, error propagation is reduced, making performance primarily dependent on the answer generation and navigation modules, which benefit from strong pretrained modules. 
The impact of these pretrained modules is further elaborated below.

\begin{figure}[t]
    \centering
    \includegraphics[width=1\linewidth]{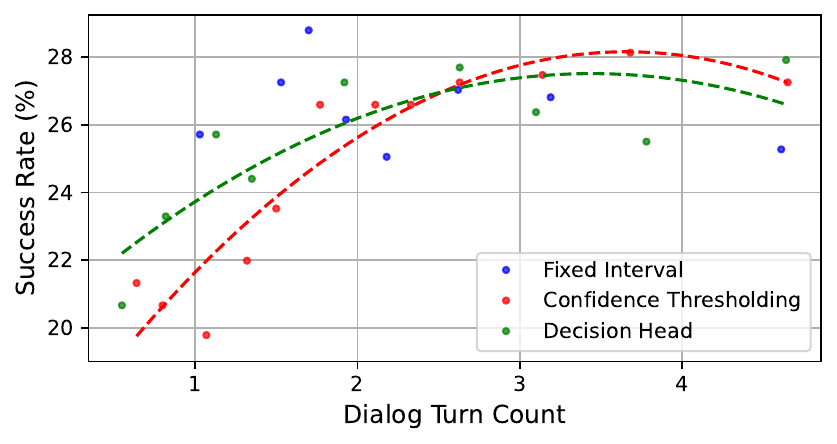}
    \vspace{-0.4cm}
    \caption{\textbf{Success Rate vs. Dialog Turn Count across three WTA strategies.} 
    We evaluate three whether-to-ask (WTA) variants; Fixed Interval (blue), Confidence Thresholding (red), and Decision Head (green), each tested with varying the interval, confidence threshold, or logit scaling, respectively.
    The red and green dotted lines is fitted curves illustrating the trend of Confidence Thresholding (red), and Decision Head (green).
    }
    \vspace{-0.5cm}
    \label{fig:wta-sr-dialog}
\end{figure}

\noindent \textbf{Effects of Pretraining} \ \ 
To evaluate the impact of pretraining on each module, we conduct an ablation study by removing pretrained weights from individual components. 
Since each module is pretrained on a related but distinct task, its contribution varies depending on the similarity between the pretraining and target tasks, as well as the dataset scale. 
The results, presented in \cref{holistic-pretrain-variation}, demonstrate the critical role of pretraining in model performance.
When the navigation module is randomly initialized (Row 2), SR and SPL drop significantly compared to Row 1, where pretrained weights are used for all components.
Additionally, both navigation and dialog efficiency decline, leading to an increased NSC and DTC.
Similarly, when the answer generation module is trained without pretrained weights (Row 5), navigation performance suffers despite a comparable NSC and DTC.
This degradation is primarily due to the frequent generation of inaccurate guidance, causing misdirections for \Nav{}.
These findings highlight the effectiveness of pretraining on relevant tasks, specifically instruction-following VLN tasks for navigation and instruction generation based on a given path for answer generation in \Task{}.
In the case of localization, the WAY~\cite{hahn2020you} dataset is relatively small, making it insufficient to provide meaningful improvements.
Pretraining the question generation also had a limited impact, likely due to misalignment between the pretraining task (instruction generation from paths) and the target task.

\begin{figure*}[ht]
    \centering
     \includegraphics[width=\linewidth]{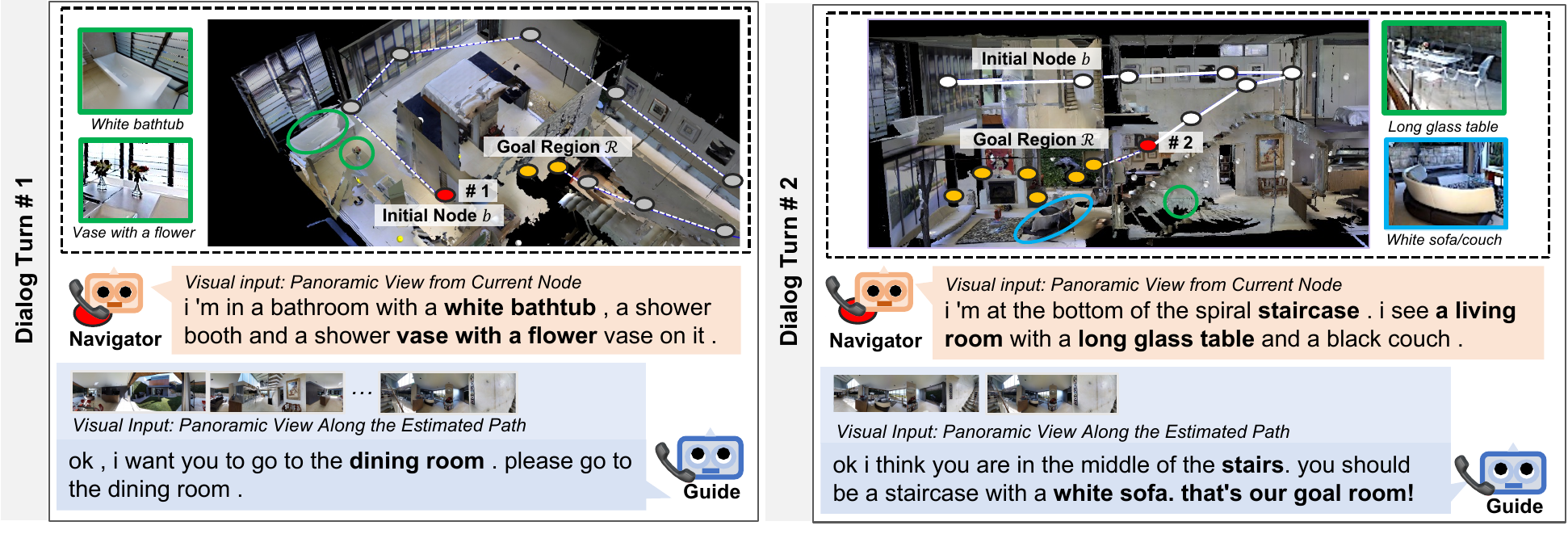}
     \vspace{-0.8cm}
    \caption{\textbf{A qualitative \Task{} episode with the initial instruction: `\textit{The goal room contains a bookcase}.'} 
    The 3D reconstructed map at the top of each turn shows the same Matterport 3D environment from different angles, with ellipses representing nodes in \Nav{}'s full path. 
    A red node indicates where \Nav{} posed a question at the current turn, yellow nodes mark the goal region, and gray and white nodes represent unvisited and visited nodes, respectively, up to the current turn.
    \Nav{} formulates questions (orange boxes) with detailed descriptions of surrounding objects (highlighted with green circles in the 3D maps and corresponding images on the side), such as `\textit{white bathtub}' in the first turn, or `\textit{long glass table}' in the second turn.
    \Gui{} provides visual hints (blue circles), like `\textit{white sofa}' in the second turn, along with path guidance (blue boxes) and instructions on where to stop (\eg, `\textit{that's our goal room}'), which are crucial for achieving high SR.
    Through this dialog, \Nav{} successfully reaches the goal region.
    }
    \label{fig:qualitative-example}
\end{figure*}
\noindent \textbf{Impact of Whether-to-Ask Strategies} \ \
We evaluate the impact of three whether-to-ask (WTA) variants by analyzing SR and DTC in \cref{fig:wta-sr-dialog}: Fixed-Interval (blue dots), Confidence Thresholding (red dots) and Decision Head (green dots).
While the Fixed Interval strategy exhibits a weak correlation between DTC and SR, both Confidence Thresholding and Decision Head show a positive correlation, enabling a trade-off between task success and dialog efficiency—an important consideration in human-in-the-loop scenarios, where excessive interaction may be perceived as intrusive. 
However, for both strategies, the benefits plateau after a few exchanges—beyond three turns for Decision Head and four for Confidence Thresholding—likely due to the modular nature of current dialog systems, which hinders context-aware interaction, and the navigation model’s limited capacity to interpret and utilize dialog content effectively.

\begin{table}
    \centering
    \scalebox{0.85}{
    \begin{tabular}{ccccccc}
    \toprule
    \textbf{Question} 
    & \textbf{Answer}
    & \textbf{SR$\uparrow$}
    & \textbf{NE$\downarrow$}
    & \textbf{LE$\downarrow$}
    & \textbf{QF$\uparrow$} 
    & \textbf{AF$\uparrow$} 
    \\
    \cmidrule(r){1-1}\cmidrule(r){2-2}\cmidrule(lr){3-7} 
    LANA~\cite{wang2023lana} & LANA~\cite{wang2023lana} & 27.0 & \textbf{11.4} & \textbf{22.6} & 4.7 & 6.8 \\
    LLaVA~\cite{liu2023llava} & LANA~\cite{wang2023lana} & \textbf{27.3} & 12.0 & 23.5 & \textbf{9.0} & 6.7 \\
    LANA~\cite{wang2023lana} & Llama~\cite{touvron2023llama2openfoundation} & 22.2 & 13.3 & 22.8 & 4.6 & \textbf{9.0} \\
    \bottomrule
    \end{tabular}
    }
    \vspace{-0.2cm}
    \caption{\label{qa-module-configuration}
    \textbf{Performance across different question and answer generation modules.}
    LANA for question or answer generation is replaced with LLaVA-1.5~\cite{liu2023llava} or Llama-3.1~\cite{touvron2023llama2openfoundation} with captions, respectively.
    \textbf{QF} and \textbf{AF} denote the fluency of questions and answers.
    LANA outperforms in NE and LE due to fine-tuning, while LLaVA and Llama generate more fluent QA owing to large-scale language training.
    \vspace{-0.5cm}
    }
\end{table}

\noindent \textbf{Dialog Fluency} \ \ 
We also assess dialog fluency, by comparing LANA, fine-tuned for the generation of questions and answers, with LLaVA-1.5 for question generation and Llama-3.1-8B for answer generation, where the Llama-3.1-8B is prompted with a set of generated captions from panoramic images along the path (\cref{qa-module-configuration}).
Fluency scores for question and answer (QF and AF) are obtained by prompting Llama-3.1-8B on a 10-point scale (details in \cref{sec:llama_evaluation} in Supp. Mat.). 
LANA scores below 5 due to limited dialog training data, producing repetitive and unstructured sentences. 
In contrast, LLaVA and Llama-3.1, trained on large-scale language data, exhibit higher fluency.
LANA outperforms LLaVA and Llama in NE and LE by generating task-specific QA. 
A key future direction is integrating LLaVA/Llama communication capabilities into navigation tasks for task-specific, environment-grounded dialog.

\noindent \textbf{Qualitative Results} \ \ 
\cref{fig:qualitative-example} presents a qualitative result of \Task{}. 
\Nav{} asks questions to clarify the path and goal location.
The generated questions incorporate key objects from the scene to aid in localization. 
\Gui{} responds with relevant visual concepts, such as room names or visible objects, either along the path or near the destination allowing successful navigation.

\section{Challenges}
\Task{} poses multiple challenges, making it particularly complex to tackle.
Below, we elaborate on these challenges.

\noindent \textbf{Expensive Data Collection} \ \ 
The interdependency among subtasks requires holistic data collection with real-time interaction between two expert annotators, making it costly and difficult to scale. 
Additionally, the turn-taking nature of the task forces one annotator to wait for the other, resulting in an inefficient and time-consuming process.
Furthermore, the limited diversity within the house simulation environment restricts scalability.
The insufficient size and diversity of the dataset result in poor generalizability in unseen environments, particularly given the task's complexity. 
Future research should investigate automatic data creation and augmentation techniques for enhanced robustness.

\noindent \textbf{Interdependancy between Subtasks} \ \ 
The interdependent subtasks complicates submodule training, as even well-performing modules in isolation may fail within the full system due to error propagation.
Minor errors in one module can cascade through subsequent stages, leading to inaccurate navigation decisions.
Thus, mitigating error propagation and ensuring robust performance across all modules is crucial for successful task execution.
Training the system as a whole could potentially mitigate these issues, but developing a unified, end-to-end model is also a significant challenge.

\noindent \textbf{Evaluation in Dynamic Context} \ \  
Evaluating \Task{} presents significant challenges due to its dynamic and non-deterministic nature. Both dialog generation and navigation involve sequential, non-definitive predictions, where slight contextual differences can significantly change the future outputs. As a result, the collected GT annotations may become invalid or unavailable for evaluation in a predicted context, further complicating system assessment. Addressing these challenges requires flexible, context-aware evaluation methods that can adapt to the evolving nature of the task.

\noindent \textbf{Long-form Multimodal Context Modeling} \ \
DialNav requires multimodal sequential modeling, processing panoramic images, action trajectories, and multi-turn dialog.
This complex dependency significantly increases task difficulty, making it challenging to integrate and utilize information effectively over extended interactions.

\noindent \textbf{Balance between Navigation and Dialog} \ \ 
Effective navigation in \Task{} requires striking a balance between task success and communication efficiency. 
Frequent interactions with \Gui{} can improve the success rate, but excessive dialog may be inefficient or disruptive, while minimizing dialog may lead to increased navigation errors.
Moreover, the agent must dynamically adapt its questioning strategy based on the reliability of \Gui{}.

\section{Conclusion}
We introduced DialNav, a dialog-based navigation task highlighting the importance of conversation due to the guide’s non-omniscient nature. To support this, we collected RAIN, a dataset of 2,231 human-human dialogs with navigation trajectories. 
We conducted experiments analyzing the current state of the art and discussed key challenges of \Task{}.
We will publicly release the dataset, code, data collection tool, and evaluation framework to foster future research.

\vspace{1em}
\noindent\textbf{\large Acknowledgements} \\
This research was supported by IITP grants
(IITP-2025-RS-2020-II201819, IITP-2025-RS-2024-00436857, RS-2024-00398115, IITP-2025-RS-2025-02263754, IITP-2025-RS-2025-02304828, No.RS-2021-II212068 (AI Innovation Hub)) and the KOCCA grant (RS-2024-00345025) funded by the Korea government
(MSIT, MOE and MSCT). Jonghyun Choi is with ECE, ASRI and IPAI in SNU.

{\small
\bibliographystyle{ieee}
\bibliography{main}
}

\clearpage
\setcounter{page}{1}
\maketitlesupplementary

\renewcommand\thesection{\Alph{section}}
\renewcommand\thefigure{\Alph{figure}}
\renewcommand\thetable{\Alph{table}}

\begin{figure*}[b]
\centering
    \begin{subfigure}{\textwidth}
        \centering
        \includegraphics[width=0.9\textwidth]{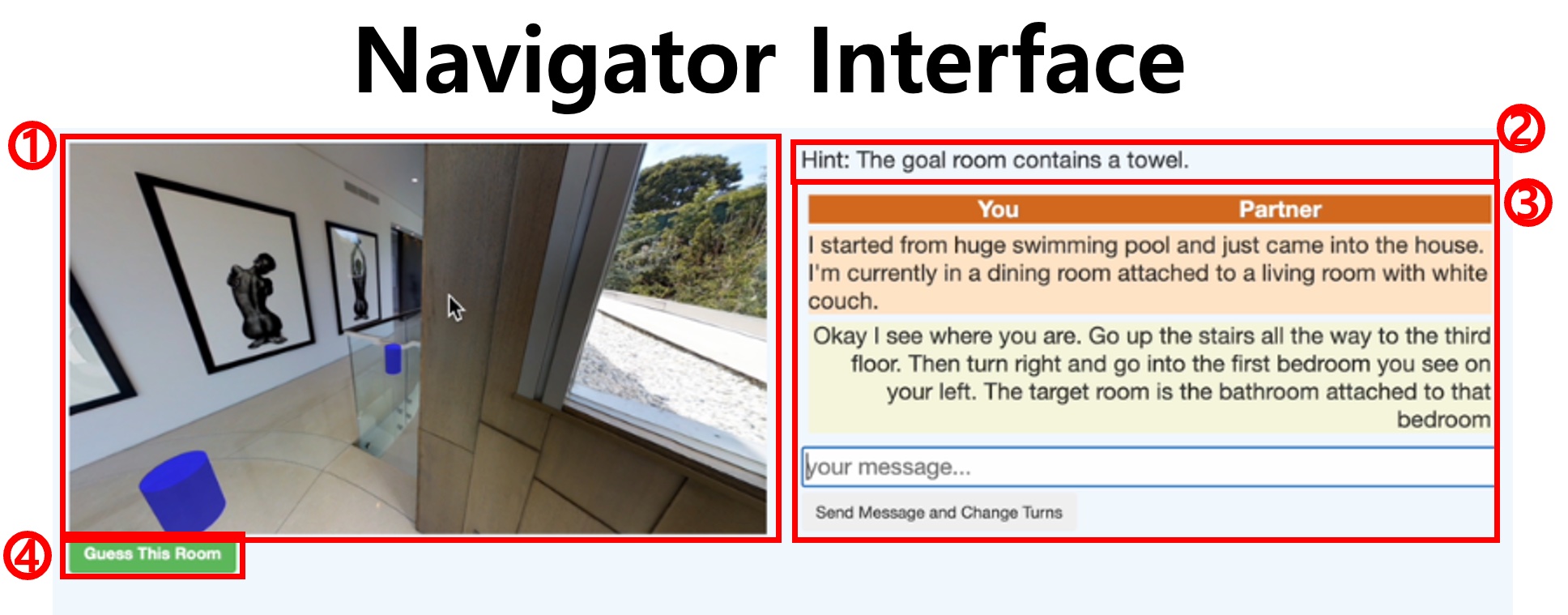}
    \end{subfigure}
    \begin{subfigure}{\textwidth}
        \centering
        \includegraphics[width=0.87\textwidth]{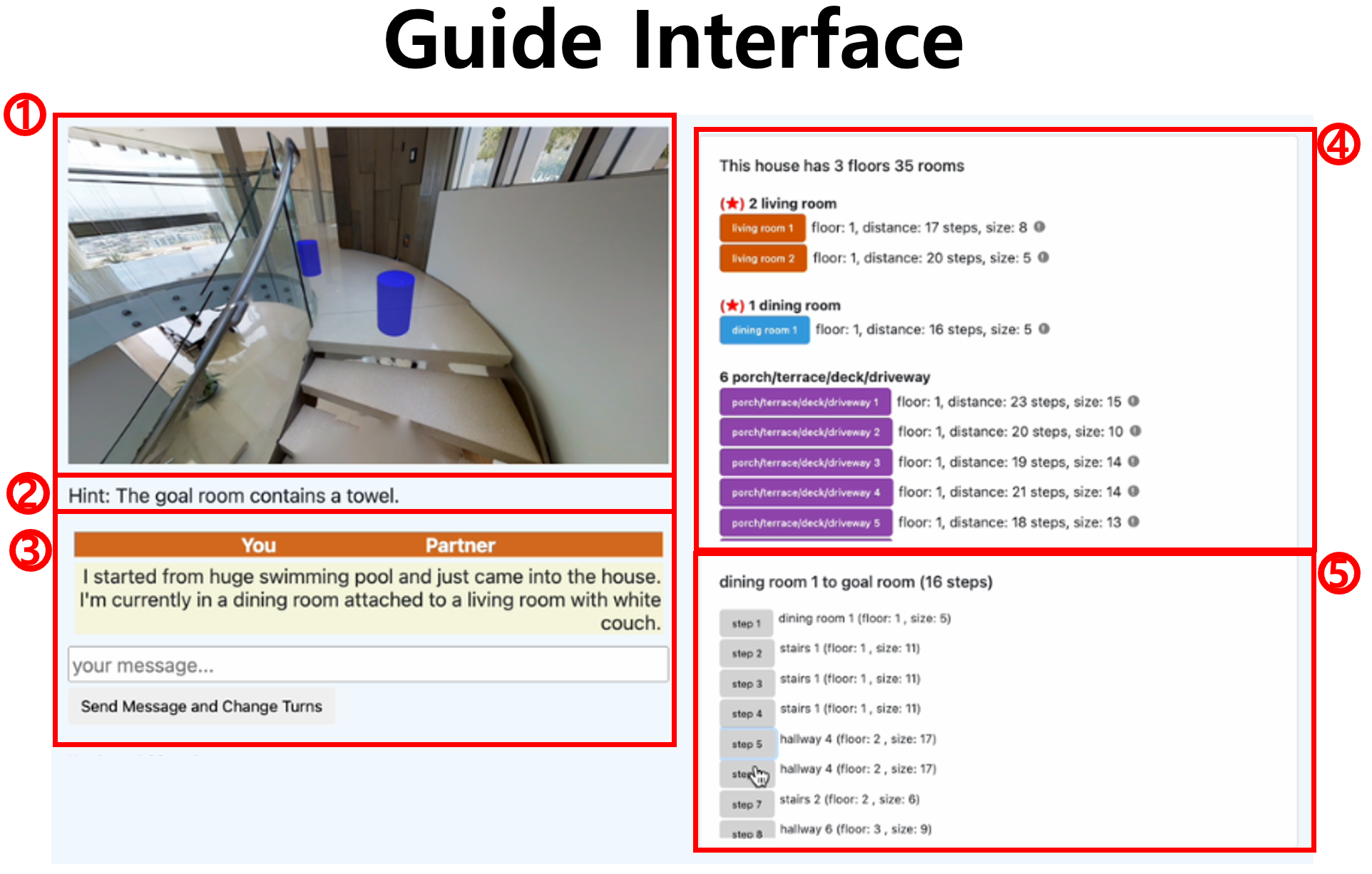}
    \end{subfigure}
\caption{\Nav{} (top) and \Gui{} (bottom) interface}
\label{fig:navigator-interface}
\end{figure*}

\section{Data Collection Interface}
\label{sec:data_collection_interface}
\cref{fig:navigator-interface} illustrates the data collection interface for \Nav{} and \Gui{}. 

\paragraph{\Nav{} Interface}
\Nav{} moves toward the goal region given the initial instruction. When unsure, \Nav{} can ask questions to receive additional information from the remote Guide.
When \Nav{} asks a question, \Nav{}'s interface is deactivated until \Gui{}'s response comes.
\Nav{} interface consists of four main components:
\begin{itemize}
    \item \textbf{Navigation Interface}: Allows free movement between nodes, active only during the Navigator's turn. 
    \item \textbf{Hint}: Provides hints about objects in the goal room, formatted as \textit{``The goal room contains \textlangle object\textrangle."}
    \item \textbf{Chat Interface}: Enables the Navigator to send questions and receive responses from the Guide.
    \item \textbf{Guess Button}: Allows the Navigator to indicate they have reached the goal. If correct, the task ends; if incorrect, a popup indicates the mistake, and the task continues. This button can be pressed multiple times without changing turns.
\end{itemize}

\paragraph{\Gui{} Interface}
\Gui{} acts as a knowledgeable participant regarding the house environment.  We designed a specialized Guide interface that simulates \Gui{} to be familiar with the entire space.  For the question of whether \Gui{} interface provided sufficient support for the task, the annotators rated 4.5 out of 5 on average.
\Gui{} interface consists of five components:
\begin{itemize}
    \item \textbf{Navigation Interface}: Allows free movement between nodes, even when it is not \Gui{}'s turn. 
    \item \textbf{Hint}: The same hint provided to \Nav{} is also shared with \Gui{}.
    \item \textbf{Chat Interface}: Enables \Gui{} to read the Navigator's questions and send responses.
    \item \textbf{House Info Interface}: Offers detailed information about the house, including:
    \begin{itemize}
        \item \textbf{Overall Information}: Presented as \textit{``This house has \(N\) floors and \(N\) rooms."}
        \item \textbf{Room List}: Displays information about all rooms grouped by type, including size, floor location, included objects, and distance to the goal region. Clicking on a room name allows \Gui{} to view and navigate nearby areas.
        \item \textbf{Shortest Path Trajectory}: The shortest path from the current location to the goal region, detailing nodes, rooms, and floors along the path. Clicking a node navigates \Gui{} to that node in the navigation interface.
    \end{itemize}
\end{itemize}

\section{Data Collection Details}
\label{sec:data_collection_details}
We recruited annotators from volunteers within the university, though participation was not limited to students. All participants were between the ages of 20 to 35.
Each participant received a compensation of 50,000 KRW for participating in the experiment for a continuous 3.5-hour session, which is equivalent to 1.45 times the minimum hourly wage.

\section{Data Preprocessing}
A total of 2,421 episodes were conducted. 
To ensure dataset quality, we first filtered out episodes that contained technical faults or exceeded the predefined time limit of 22 minutes. 
Additionally, we manually reviewed outlier episodes based on three criteria: trajectory detour, dialog count, and annotator scores. 
Specifically, we identified episodes within the top 1\% of outliers for each metric—5.6 for detour, 6 for dialog count, and 1 for annotator scores. 
Upon review, these episodes were retained, as they were deemed sufficiently valid for the task. 
\cref{tab:sample-dialnav-dialogs} and \cref{tab:outlier-dialogs} presents examples of qualitative DialNav samples and outliers respectively.
To further refine the dataset, we manually corrected instances where dialog entries were inadvertently split due to user errors, such as accidental key presses. 
After these refinements, the final dataset consisted of 2,231 episodes. The dataset was then partitioned into training, validation, and test splits, with 1,559 episodes for training, 285 for testing, and 387 for validation. 
The validation set was further divided into seen (111 episodes) and unseen (276 episodes) environments, while the test set exclusively contained unseen environments. 
We adhered to the seen/unseen/test environment split established in prior work \cite{anderson2018vision}.

\section{The \InstanceData{} Dataset}
A \Dataset{} episode consists of multiple dialog turns and corresponding actions.
For training, we restructured episodes into segment instances following the methodology of CVDN~\cite{das2018embodied}.
Specifically, an episode with three dialog turns generates four segment instances: one representing the trajectory before any dialog and three corresponding to successive dialog turns.

Each \InstanceData{} is defined as ($G$, $I$, $\mathcal{R}$, $\mathcal{T}_t$, $\mathcal{D}_i$), where $G$, $I$, and $\mathcal{R}$ are common across the entire episode, while $\mathcal{D}_i$ is dialog history until dialog turn $i$ and $\mathcal{T}_t$ is trajectory history up to dialog turn $i$.
Each \InstanceData{} instance represents a specific state within an episode, capturing the interactions between the \Nav{} and the \Gui{}.

For Navigation training, all instances are utilized, while for Localization and QA generation, first segment in each episode which does not contain dialog is not included.
We adhered to the seen/unseen/test environment split established in prior work \cite{anderson2018vision}.
Refer to \cref{tab:dataset-splits} for the respective counts.

\begin{table}[h]
    \centering
    \small
    \begin{tabular}{c|cccc|c}
    \hline
    \textbf{Type} & \textbf{Train} & \textbf{ValSeen} & \textbf{ValUnseen} & \textbf{Test} & \textbf{Total} \\
    \hline
    DialNav & 1,559 & 111 & 276 & 285 & 2,231 \\ 
    Instance & 4,493 & 337 & 805 & 768 & 6,403 \\ 
    w/ dialog & 2,934 & 226 & 529 & 483 & 4,172 \\ 
    \hline
    \end{tabular}
    \caption{Dataset size per splits}
    \label{tab:dataset-splits}
\end{table}

\begin{figure*}
    \centering
    \small
    \begin{tabular}{|cp{0.8\linewidth}|}
        \hline
        \Tstrut Q: & Hi I'm in a grey bedroom. The bed has dark brown frame and grey patterned sheets. There's a grey sofa at the foot of the bed, facing a TV. I can see two big windows. \\
        \Tstrut A: & Please exit the room using the door on the right. Then walk down the hallway and you'll find a kitchen on your left and a living room with sofas on your right. Go through them and you'll see pictures of a couple and their child on the wall. There are 6 in total with different shapes. Do you see them?  \\
        \Tstrut Q: & Yes. I'm standing in front of them. \\
        \Tstrut A: & Ok now take a right turn and walk a few steps ahead. Then when you see two pictures of a boy and a girl on the wall, make a right turn and you'll see a bedroom. Our destination is the bathroom attached to the bedroom 
        \Bstrut \\
        \hline
    \end{tabular}
    \vspace{0.05cm}

    \begin{tabular}{|cp{0.8\linewidth}|}
        \hline
        \Tstrut Q: & I was standing in the corner, which has two ways to go. There are lots of rooms and they look all same. I entered a nearest room, and I can see a white table and a black chair. On the door, number '321' is written. \\
        \Tstrut A: & Great. Go out of the room and turn left. Walk straight down the hallway up until you see 4 red sofas on your left. Tell me when you get there \\ 
        \Tstrut Q: & I can see the red sofas. I am facing them. \\
        \Tstrut A: & Okay. Turn right there. Having the red sofas on you back, walk down the hallway up until you see a white pillar and a poster of a men attached to the glass wall on you left. And a exit sign on your right \\
        \Tstrut Q: & I am there! Facing the exit sign. \\
        \Tstrut A: & Great! Here's the tricky part. Walk more down the hallway, until you find a messy room with a mircrowave on your right, and in front of the goal room, there's another small office with only one white desk and chair.
        \Bstrut \\
        \hline
    \end{tabular}
    \vspace{0.05cm}

    \begin{tabular}{|cp{0.8\linewidth}|}
        \hline
        \Tstrut Q: & Hello, I am in a small bathroom with a toilet and a black sink. There is a door to my right which is wooden brown, and also another door to my left which is white. \\
        \Tstrut A: & Alright, go through the white door, and go straight at the sofa. If you stand in front of a table with fruits, that's the goal room.
        \Bstrut \\
        \hline
    \end{tabular}
    \vspace{0.05cm}

    \begin{tabular}{|cp{0.8\linewidth}|}
        \hline
        \Tstrut Q: & now im in a bedroom; a bed with patterned bedding, an abstract painting of two horses right above the bed, a black chair set in front of the bed. \\
        \Tstrut A: & You're on the 4th floor and we're gonna have to make a way to the 2nd floor. So I want you to come out of the bedroom through the door on your right: to the TV side. Once you're out, there will be a lounge with a round wooden table and a staircase going down on your 2 o'clock.  We'll take down those round stairs and let me know when you're down and find another stairs in front of you. \\
        \Tstrut Q: & yep now im facing another stairs \\
        \Tstrut A: & Yes head down the stairs to the entry lobby and make a U-turn to your left and there will be another short stairs for you to go down. Once you're down again from those stairs, go to the living room that you can find on your 2 o'clock: right side of the statue at your front. Once you've made your way to the living room, take a right turn and head to the pool table. There's a bathroom on the right side of that pool table and that should be our goal room.
        \Bstrut \\
        \hline
    \end{tabular}
    \vspace{0.05cm}

    \begin{tabular}{|cp{0.8\linewidth}|}
        \hline
        \Tstrut Q: & I'm in a very fancy living room with a greyish blue fireplace with a mirror above it. There is also a chandelier, and the pattern on the floor has some red in it. Should I leave the room into a long hallway? \\
        \Tstrut A: & There are several rooms that have a greyish blue fireplace, so could you please elaborate on it more? For example, it has a brown wooden round table, it has a white striped patterned sofas, like that. Thanks. \\
        \Tstrut Q: & Oh, sorry. It has a long white couch, with 5 small white chairs aligned next to it. There is also a small bed in the corner. \\
        \Tstrut A: & Oh, I found you. Enter the hallway, which is on the left side of the fireplace. If you walk through the hallway, you'll see the stairs on your right side. Could you go halfway down the stairs? \\
        \Tstrut Q: & Yes, I'm halfway down the stairs. Keep going? \\
        \Tstrut A: & You have to stop. Is the stair you've walked down on your behind? \\
        \Tstrut Q: & Yes, and there are stairs on either side behind me. I'm facing the white wall. \\
        \Tstrut A: & Great. Now just take one step at the stair in right side of you. That is the goal room.
        \Bstrut \\
        \hline
    \end{tabular}
    \vspace{0.05cm}
    \vspace{0cm}

    \caption{\textbf{Sample RAIN dialogs}}
    \label{tab:sample-dialnav-dialogs}
\end{figure*}
\begin{figure*}
    \centering
    \small
    \begin{tabular}{|cp{0.8\linewidth}|}
        \hline
        \Tstrut Q: & I'm in a room with a swimming pool. There are 3 sunbeds, and I see a treadmill on one side. \\
        \Tstrut A: & good new. it's near. assuming you are facing the swimming pool, turn right and take a few step and turn right and you will see two spa rooms: take the one on the left and the room has this fan with red flowers and a blanket with check patterns. Text me if you aren't confused!  \\
        \Tstrut Q: & I'm looking at the swimming pool, and there are sunbeds behind me. If I turn right and another right, I see a concierge, not spa rooms. \\
        \Tstrut A: & ok. there are two pools right? Stand in front of the larger pool. (The one on the left) and you see that there are plants next to that pool. There are two rooms behind those plants. I need you to go behind that plant, so I need you to head towards the diamond shape wall. From here, turn right. Let me know if you find the diamond wall.
        \Bstrut \\
        \hline
    \end{tabular}
    \vspace{0.05cm}
    \begin{tabular}{|cp{0.8\linewidth}|}
        \hline
        \Tstrut Q: & Hi. I'm in the room with a huge bed which has a purple bedding, and I can also see the light green colored sofa. I guess I'm at the second floor. \\
        \Tstrut A: & Hi. Is the roof inclined not flat? And do you see trees through a window? \\ 
        \Tstrut Q: & Yes. The roof is inclined, and I can see tress through the window. \\
        \Tstrut A: & Great! You're currently on the third floor. First come out from the bedroom. While you come out from that room, trees should be on your left side. \\
        \Tstrut Q: & Okay. Should I go down stairs? \\
        \Tstrut A: & Yes. Please go down the stairs. \\
        \Tstrut Q: & I came down. Now I'm standing at the kitchen, and can see a big wood table.\\
        \Tstrut A: & Terrific! You will see a hallway at the left side of that big wood table. Please go to the hallway. \\
        \Tstrut Q: & I'm at the hallway now. I can see a pool through the window. \\
        \Tstrut A: & Sorrry! Do you see a black car through a window? \\
        \Tstrut Q: & I guess I was at the different hallway. I got to the other one, and now I can see the black car through the window. \\
        \Tstrut A: & Fantastic! There are brown stairs between white walls. Please go down to the first floor by following those stairs. \\    
        \Tstrut Q: & I came down! Now I'm looking ahead a room with work out machines. \\
        \Tstrut A: & Good. Please do not go into that room. Instead enter a room next to the room with workout machines. Then you will see a wall on which a kind of graffiti is drawn. \\
        \Tstrut Q: & I'm in! \\
        \Tstrut A: & The place in front of that wall. That is the goal.
        \Bstrut \\
        \hline
    \end{tabular}
    \vspace{0.05cm}
    \begin{tabular}{|cp{0.8\linewidth}|}
        \hline
        \Tstrut Q: & in front of stairs going down, on my left on the wall is an abstract painting, signiture reads what looks like 'Hunk' \\
        \Tstrut A: & i got it. it's quite complicated path to our room. putting that painting on your right, please make a right turn.(kind of u turn you should make) then you might see set of wooden stairs leading to upstairs. going up, you can make a right turn to see a few black stairs. finishing that stair, there might be a kitchen. please make a right turn right at the person like-statue. then there might be a set of wooden stairs, and please go upstairs.finisnihing that, you might find a bed room and going inside, there is a spa. that's our goal room. \\ 
        \Tstrut Q: & there's no right turn I can make if I put the painting on my right. Im in front of stairs going down, behind me is an open white door (a room with two big yellow sofa beds) \\
        \Tstrut A: & oh isn't that the yellow painting? \\ 
        \Tstrut Q: & yes, if i make a right turn, i'm in a sauna. just tell me which floor I need to be on. do i need to go downstairs? \\
        \Tstrut A: & our goal room is on the 3rd floor. if that's wooden spa, i guess its our goal room  
        \Bstrut \\
        \hline
    \end{tabular}
    \vspace{0.05cm}
    \begin{tabular}{|cp{0.8\linewidth}|}
        \hline
        \Tstrut Q: & I'm at front of stairs, do I have to go down? \\
        \Tstrut A: & There are 3 stairs in this house. I'll need more infos. Please give me more details.\\ 
        \Tstrut Q: & I can see two bedrooms, one has grey blanket on the bed, and the other one has brown blanket on the bed.
 \\
        \Tstrut A: & So, you're in hallway 1, floor2. Our goal room is bedroom 4. Please enter bedroom with the brown blanket. That's our goal room.
        \Bstrut \\
        \hline
    \end{tabular}
    
\caption{\textbf{Sample dialog with outliers}:
(1) Dialog with a significant detour,  
(2) Dialog with a high number of QA turns,  
(3) Dialog with a low Guide score,  
(4) Dialog with a low Navigation score.}
\label{tab:outlier-dialogs}
\end{figure*}

\section{Module-wise Training and  Experiments}
\label{sec:module-wise_training}
\renewcommand{\arraystretch}{1.2}
\begin{table}
\centering
\scalebox{0.9}{
    \begin{tabular}{lcccc}
    \toprule
    \textbf{Method} &\textbf{+SV}& \textbf{Val Seen}
    & \textbf{Val Unseen}
    & \textbf{Test}
    \\
    \midrule
    Shortest
    &&  22.51  & 23.06  & 25.23  \\
    Random
    & & 1.91  & 1.94  & 0.25 \\
    \arrayrulecolor{gray}\cdashline{0-4}\arrayrulecolor{black}
    HAMT~\cite{chen2021history}
    & & 11.26 & 8.31  & 5.87 \\
    DUET~\cite{chen2022think}
    & & 11.13 & 10.22& 11.53 \\
    DUET~\cite{chen2022think}
    & \checkmark &  \textbf{12.88} & \textbf{12.07} & \textbf{12.37} \\
    \bottomrule
    \end{tabular}
}
\caption{\label{navigation-scores}
\textbf{Navigation scores on \InstanceData{}.}
The goal progress of navigation to the destination is measured when the initial instruction and the last answer from the previous dialogue are given. \textbf{+SV}: Pretrained on ScaleVLN~\cite{wang2023scaling}.
}
\end{table}

\paragraph{Navigation.}
We utilized two existing VLN models for navigation: HAMT~\cite{chen2021history} and DUET~\cite{chen2022think}.
HAMT is first initialized with publicly available pretrained weights, and finetuned on \Task{} training instances. 
To evaluate the impact of the large-scale pretraining on \Task{} navigation, we compared the DUET with weights pretrained on ScaleVLN~\cite{wang2023scaling}.
For each \InstanceData{} instances, the initial instruction and the dialog history is appended to form a navigation instruction for each instance. 
The VLN models were trained to determine the best next optimal action given the instruction. 
Due to the multi-turn nature of the dialog, the answers often do not contain the complete path to the goal. 
Therefore, Goal Progress (GP)~\cite{thomason2020vision}, the distance that agent got closer towards the goal, is used as navigation metric.

\cref{navigation-scores} shows the navigation performance on \InstanceData{}.
We report shortest path length and random agent performance serving as an upper and lower bounds respectively.
The DUET~\cite{chen2022think} model, which retains past trajectory history as a graph, demonstrated superior performance in unseen environemnt compared to the HAMT~\cite{chen2021history} model. 
Additionally, pretraining on ScaleVLN~\cite{wang2023scaling} further improved navigation performance across all environment splits.
This result indicates the advanced model and large-scale pretraining on VLN models brings high performance in \Task{} navigation as well.

Through these experiments, we selected DUET~\cite{chen2022think} pretrained with ScaleVLN~\cite{wang2023scaling} which showed the best performance across all environments for our baseline model.

\begin{table}[t]
\centering
\scalebox{0.85}{
\begin{tabular}{lcccccccccc} 
\toprule
&\multicolumn{2}{c}{\textbf{Val Seen}} & \multicolumn{2}{c}{\textbf{Val Unseen}} & \multicolumn{2}{c}{\textbf{Test}} \\
\textbf{Method} & \textbf{B4 \(\uparrow\)}
& \textbf{RG \(\uparrow\)} 
& \textbf{B4 \(\uparrow\)}  
& \textbf{RG \(\uparrow\)} 
& \textbf{B4 \(\uparrow\)} 
& \textbf{RG \(\uparrow\)} 
\\
\cmidrule(r){1-1} \cmidrule(lr){2-3} \cmidrule(lr){4-5} \cmidrule(lr){6-7}
\textbf{LLaVA~\cite{liu2023llava}} & .0311&.2074& .0318&.2183& .0259&.2043\\
\textbf{LANA~\cite{wang2023lana}} &.0405&\textbf{.2087}&\textbf{.0533} & \textbf{.2203}&.0502 & \textbf{.2131} \\
\textbf{LANA(pt)~\cite{wang2023lana}} &\textbf{.0532}&.2005&.0527 & .2111&\textbf{.0539}&.2071\\
\bottomrule
\end{tabular}
}
\vspace{-0.2cm}
\caption{
\textbf{Question Generation scores on \InstanceData{}.}
The similarity between the questions generated by the model and those generated by humans in the same context is measured.
\textbf{B4}: BLEU4, \textbf{RG}: Rouge-L}
\vspace{-0.3cm}
\label{question-generation-scores}
\end{table}

\paragraph{Question Generation.}
We employed the VLN instruction generation model LANA~\cite{wang2023lana} and the multimodal model LLAVA-1.5~\cite{liu2023llava} for generating questions. 
We tested LANA with and without pretraining . 
Although \Nav{} has access to the past trajectory and previous dialogs, to simplifying the task, we provided only the visual input from the current viewpoint in this work.
LLAVA-1.5 (7B) is simply prompted to generate question including details of current panoramic view. 

The objective of Question Generation is to facilitate human-like dialog; therefore, we evaluated the similarity of the generated questions to those asked by humans in identical situations.
\cref{question-generation-scores}, reports metrics for question generation.
As discussed in the main manuscript, the improvement by pretraining LANA is limited in question generation due to the significant mismatch between pretext and target tasks. 

Through these experiments, we selected LANA~\cite{wang2023lana} with pretrained weight for our baseline model.

\paragraph{Whether to Ask.}
Whether to Ask (WTA) is a task to predict binary decision of to ask or not given navigation context. 
Previous dialog-based studies have used simple Fixed-Interval~\cite{roman2020rmm}, Confidence Thresholding~\cite{zhu2021self,gao2022dialfred} or adopting an additional model~\cite{singh2022ask4help}.
We tested all these 3 approaches. 
In Confidence Thresholding, questions are triggered when action confidence falls below a threshold. 
For the third approach, we added a decision head to DUET~\cite{chen2022think} model to utilize the intermediate output of the action decision and trained with \InstanceData{}.
We selected the decision head for our baseline model.

\begin{table}[t]
    \centering
    \scalebox{0.9}{
    \begin{tabular}{l ccc cc cc} 
    \toprule
    &\multicolumn{3}{c}{\textbf{Language}} & \multicolumn{2}{c}{\textbf{Navigation}} \\
    \textbf{Method} & \textbf{B4 \(\uparrow\)} & \textbf{RG \(\uparrow\)} & \textbf{CD \(\uparrow\)} & \textbf{GP\(\uparrow\)} & \textbf{SR\(\uparrow\)}\\
    \cmidrule(r){1-1} \cmidrule(lr){2-4} \cmidrule(lr){5-6}
    \textit{Val Seen} & & & &
    \\
    \textbf{Llama-3.1} & .0268 & .1802 & .0595 & 5.36 & 18.92 \\
    \textbf{LANA} & .0539 & .2156 & .0953 & 6.89 & 23.42\\
    \textbf{LANA(pt)} & \textbf{.0745} & \textbf{.2342} & \textbf{.1310} &\textbf{10.63}&\textbf{30.09} \\
    \midrule
   \textit{Val Unseen} & & & &
   \\
    \textbf{Llama-3.1} & .0271 & .1765 & .0394 & 5.35& 13.01 \\
    \textbf{LANA} & .0542 & .2082 & .0722 & 8.03 & 22.52\\
    \textbf{LANA(pt)} & \textbf{.0596} & \textbf{.2223} & \textbf{.1086} & \textbf{8.92} & \textbf{25.28} \\
    \midrule
    \textit{Test} & & & & 
    \\
    \textbf{Llama-3.1} & .0280 & .1881 & .0550 & 6.74 & 13.08 \\
    \textbf{LANA} & .0519 & .2217 & .0789 & 6.93 & 10.55\\
    \textbf{LANA(pt)} & \textbf{.0648} & \textbf{.2325} & \textbf{.0952} &  \textbf{9.50} &  \textbf{17.11} \\
    \bottomrule
    \end{tabular}
    }
    \caption{\label{table-answer-generation-module}
    \textbf{Answer generation scores.} 
    \textbf{LS:} Llama-S, \textbf{LF:} Llama-F, \textbf{GP (U):} GP on unseen validation set, \textbf{GP (S):} GP on seen validation set. 
    The navigation scores (GP, SR) report the performance of the navigation task based on the answers generated by the model, using the DUET model pretrained with ScaleVLN.
    }
    \end{table}

\begin{table}[t] 
\centering
\footnotesize
\begin{tabular}{lcccccc} 
\toprule
&\multicolumn{2}{c}{\textbf{Val Seen}} & \multicolumn{2}{c}{\textbf{Val Unseen}} & \multicolumn{2}{c}{\textbf{Test}} \\
\textbf{Method} & \textbf{LE\(\downarrow\)} & \textbf{A@3\(\uparrow\)} & \textbf{LE\(\downarrow\)} & \textbf{A@3\(\uparrow\)} & \textbf{LE\(\downarrow\)} & \textbf{A@3\(\uparrow\)} \\
\cmidrule(r){1-1} \cmidrule(lr){2-3} \cmidrule(lr){4-5} \cmidrule(lr){6-7}
Random 
& 19.76 & 7.96 & 18.77 & 4.34 & 21.54 & 3.46 \\
\arrayrulecolor{gray}\cdashline{1-7}\arrayrulecolor{black}
SCN& 
\textbf{10.20} & \textbf{44.69} & 14.63 & 23.58 & 16.04 & 19.14 \\
SCN†& 
10.97&42.04&13.87&23.58&15.60&22.20 \\
GCN& 
12.09&38.94&13.33&24.91&16.53&16.90 \\
GCN†& 
11.47 & 36.73& \textbf{11.63} & \textbf{31.19} & \textbf{15.28} & \textbf{22.77}\\ 
\bottomrule
\end{tabular}
\vspace{-0.2cm}
\caption{
\textbf{Localization scores on \InstanceData{}.} 
Scores are reported with the model with least LE for in Validation Unseen set in each model.
\textbf{LE}: Localization error in meters.
\textbf{A@3}: Accuracy with an allowable error margin of 3m. 
}
\vspace{-0.3cm}

\label{table-localization-module}
\end{table}

\paragraph{Localization.}
Localization task predicts location of \Nav{} given the previous dialog.
Although \Gui{} has access to the full dialog history, we only use the last question in dialog for this work for simplification. 
We evaluate two ranking models from~\cite{hahn2020you}: Simple Cross-Modal Network (SCN) and Graph Convolutional Network (GCN).
Both compute node embeddings for the environment graph $G$ and obtain a cross-modal feature by element-wise multiplying it with the query embedding, followed by a linear layer for ranking.
The key difference is that SCN models each node independently, whereas GCN captures graph structure.
All models are first pretrained on the WAY~\cite{hahn2020you} dataset and then finetuned on \InstanceData{}.
For evaluation, we adhere to the protocols in \cite{hahn2020you,hahn2022transformer}, measuring localization error in meters and the accuracy of successful localization within 3 meters.

The SCN exhibits a significant performance disparity between seen and unseen environments, with minimal performance gains from WAY pretraining. (\cref{table-localization-module} ln2, 3)
In contrast, the GCN model demonstrates greater gains from pretraining and better generalization to unseen environments.  (\cref{table-localization-module} ln4, 5) 
This suggests that more data and complex models are required to generalize effectively to novel environments. 
The localization task remains underexplored, necessitating further research.

Through these experiments, we selected GCN with pretrained weight for our baseline model.

\paragraph{Answer Generation.}
We tested VLN instruction generation model LANA with and without pretraining. We also tested Llama-3.1-8B-instruct~\cite{touvron2023llama2openfoundation}
 with promppting to test power of large language model for answer generation task. 
For Llama, we first generates detailed caption of each navigation nodes with LLaVA-1.5~\cite{liu2023llava} and prompts Llama to generate navigation instruction based on the sequence of captions through route. 
See ~\cref{fig:llava_caption} and ~\cref{fig:llama_answer} for LLaVA and Llama prompts. 
Although \Gui{} has full access to previous dialog and entire house map, we only use the remaining trajectory to the goal as input to simplified the task.

The objective of Question Generation is to facilitate human-like dialog; therefore, we evaluated the similarity of the generated answers in identical situations. 
Answer generation is similar to the VLN instruction generation task, so metrics improved through pretraining.
To check the impact on navigation, we also reported goal progress and SR of DUET~\cite{chen2022think} model pretrained with ScaleVLN~\cite{wang2023scaling} when the answers were given as instructions. 

Through these experiments, we selected LANA~\cite{wang2023lana} with pretrained weight for our baseline model.

\section{Question Generation Examples}

\begin{figure}[ht]
\centering
\includegraphics[width=\linewidth]{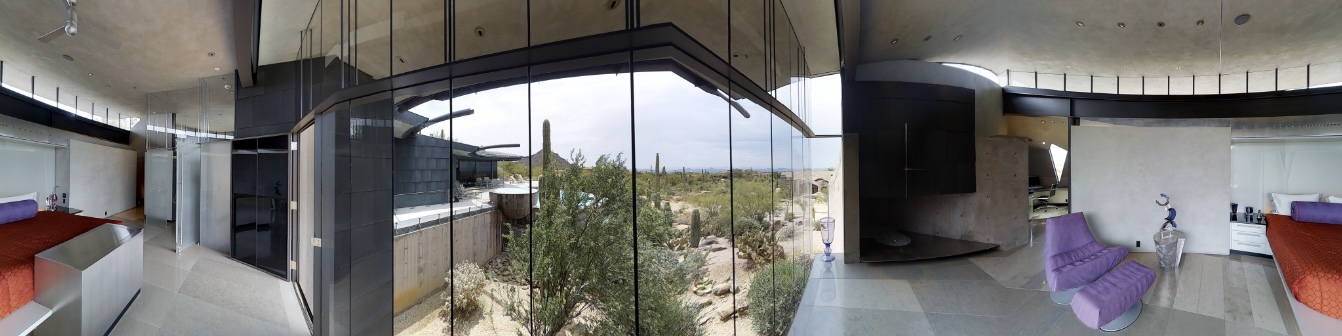} \\
\begin{tabular}{p{0.95\linewidth}}
\textbf{Ground Truth (GT):} I'm at a bedroom. There is a purple sofa. The bed has red blankets, one purple cushion and two white pillows. The outside is clearly showing, with trees and multiple cactus. \\
\textbf{LLaVA:} I'm in a room with a large window, and I can see a view of a desert landscape outside. \\
\textbf{LANA:} I'm in a hallway with a large glass window. Next to the sink is a glass shelf to the outside. There is a glass door to the right next to the sink. To the sink, there is a glass shelf to the kitchen. \\
\end{tabular} \\

\caption{Qualitative analysis of question generation from different models.}
\label{fig:question-generation-example}
\end{figure}
\cref{fig:question-generation-example} shows example of question generation given a viewpoint.

\section{Answer Generation Examples}
\begin{figure}[ht]
\centering

\includegraphics[width=\linewidth]
{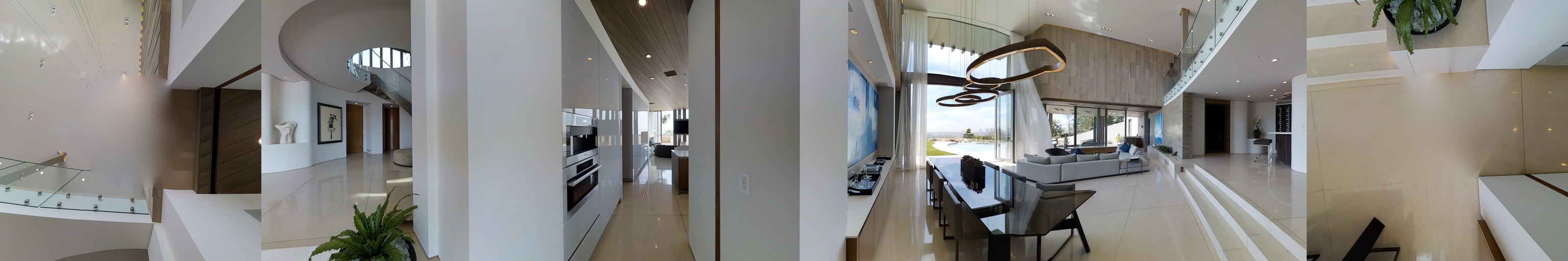} \\
\includegraphics[width=\linewidth]{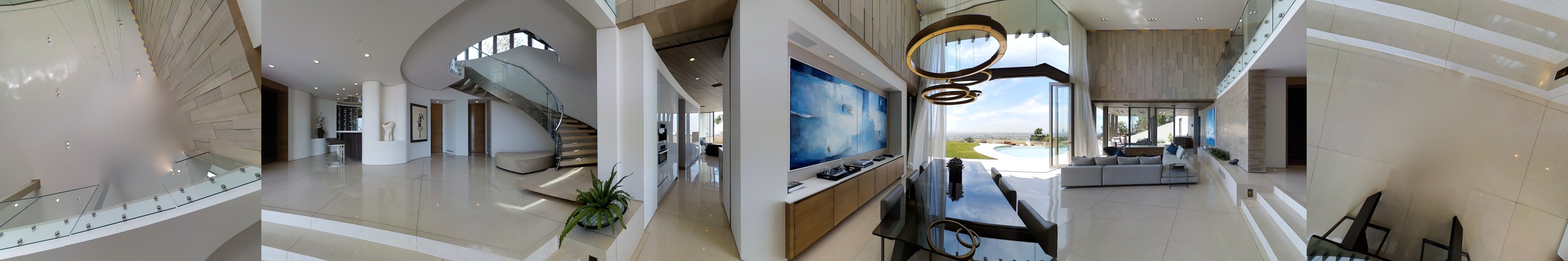} \\
\includegraphics[width=\linewidth]{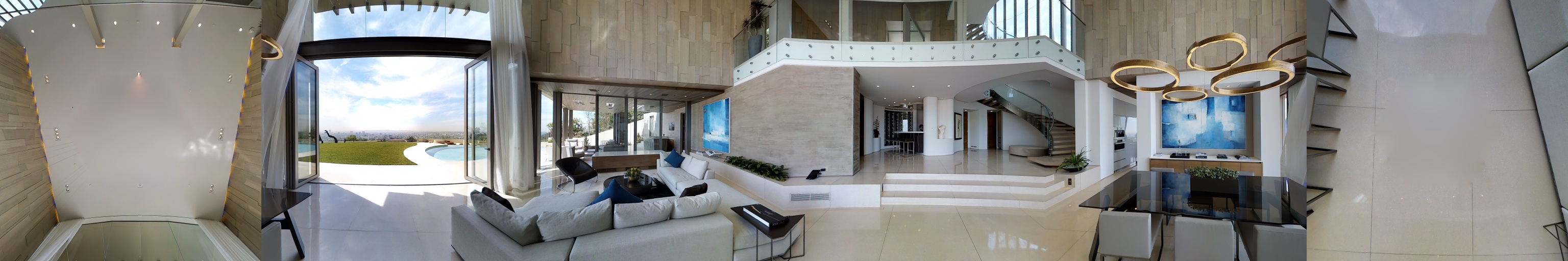} \\
\includegraphics[width=\linewidth]{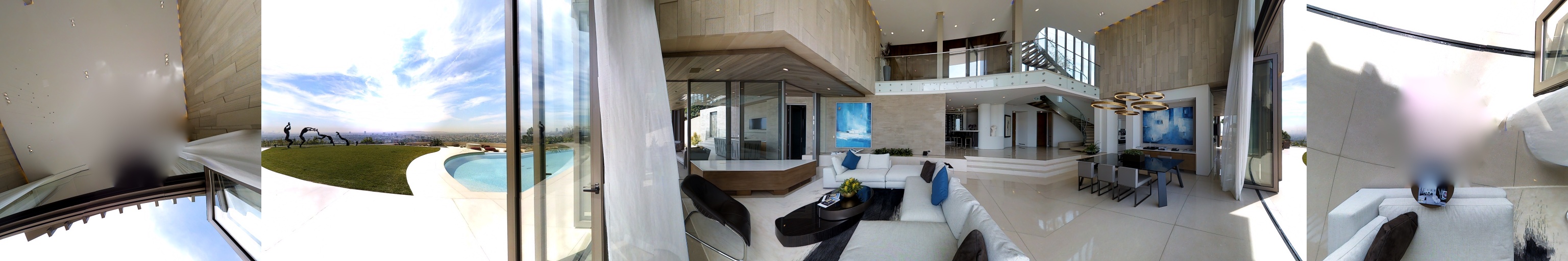} \\
\includegraphics[width=\linewidth]{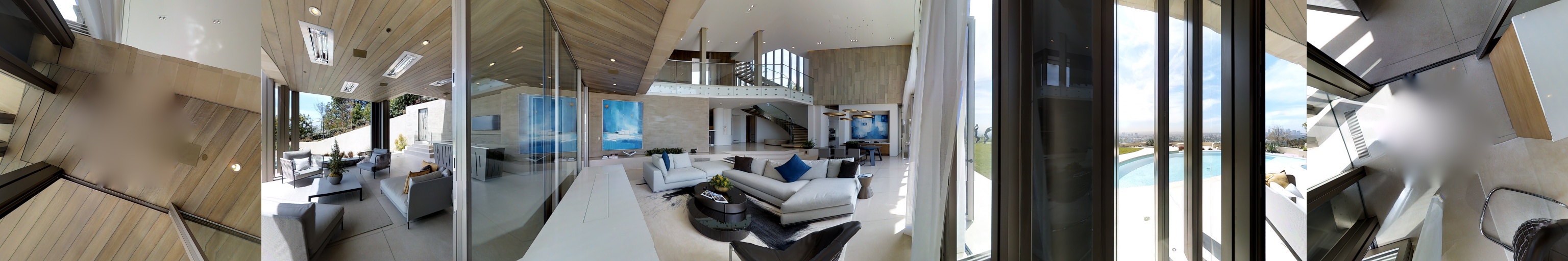} \\
\includegraphics[width=\linewidth]{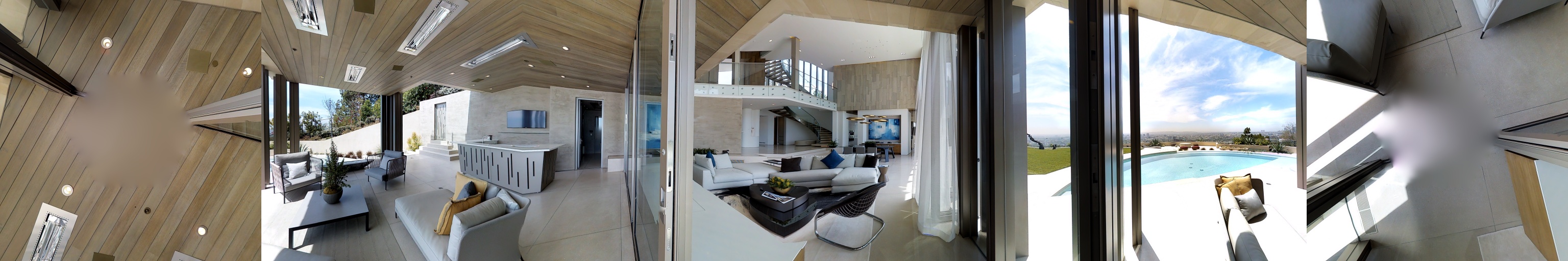} \\
\begin{tabular}{p{0.95\linewidth}}
\textbf{Ground Truth (GT):} Almost there. Now walk towards the light grey sofa in front of you. On the left to that sofa, there is a path to a terrace where you can see the pool on the left. The terrace is our target
 \\
\textbf{LLaVA:} Head towards the large window with the modern design, and you'll find the dining area with a table and chairs. The kitchen is adjacent to the dining area, and you can access it through a glass door. If you'd like to explore more, the staircase leading to another floor is located in the living room, which is decorated with a vase, a TV, and several books. \\
\textbf{LANA:} great . so , facing the white couch , you will see a grey couch on your right , and a tv on your right . go straight until you see a living room with a tv on your right . that 's our goal room . 
 \\
\end{tabular} \\
\caption{Qualitative analysis of answer generation from different models.}
\label{fig:answer-generation-example}
\end{figure}

\cref{fig:answer-generation-example} shows an example of answer generation based on a list of panoramic views along the route to the goal room.

\section{LLaVA Prompt for Question Generation}
\begin{figure}[t]
    \centering
    \setlength{\fboxsep}{8pt}
    \fbox{%
        \parbox{0.4\textwidth}{
            \small
            \textbf{Prompt:} \\
            \texttt{You will be given a panoramic image of indoor scene. Create a sentence including what type of room the given image is (ex. bedroom, bathroom, empty room, stair, hallway), and details and any unique objects that would not appear in other regions of the building so that someone else can easily locate. Start with `I'm in' or `I can see'.} \\[8pt]
            \textbf{LLaVA Output:} \\
            I'm in a room with a wooden door and a blue curtain.
        }
    }
    \caption{\textbf{Prompt for LLaVA's question generation and its results.}}
    \label{fig:llava_prompt}
\end{figure}
The prompt for LLaVA-1.5 used in question generation is shown in \cref{fig:llava_prompt}. We provide LLaVA with a panoramic image to generate a scene description as a \Nav{} question.

\section{Llama based Answer Generation}
\subsection{Generating LLaVA captions}
\begin{figure}[t]
    \centering
    \setlength{\fboxsep}{8pt}
    \fbox{
        \parbox{0.4\textwidth}{
            \small
            \textbf{Prompt:} \\
            \texttt{Please describe the {region} in this image.} \\[8pt]
            \textbf{LLaVA Output:} \\
            The bathroom in the image is a small space with a toilet and a sink. The sink is located near the toilet, and there is a book placed on the countertop. The bathroom appears to be part of a larger living area, as it is situated next to a wall and a door.
        }
    }
    \caption{\textbf{Prompt for LLaVA's caption generation and its results.}}
    \label{fig:llava_caption}
\end{figure}

The prompt for LLaVA-1.5 caption generation is shown in \cref{fig:llava_caption}. We provide LLaVA with a list of panoramic images and extract captions for the images corresponding to the path.
\subsection{Prompting Llama to generate answers}
\begin{figure}[t]
    \centering
    \setlength{\fboxsep}{8pt}
    \fbox{
        \parbox{0.4\textwidth}{
            \small
            \textbf{Prompt:} \\
            \texttt{You are an agent for creating navigation route. Given sequence of scene image description, create a navigation guide sentence for the route. You don't have to describe every single step. Try to add any unique object or landmark. Return your evaluation results in the following JSON format without any additional text: \\
            {`response': `<your response>'}} \\[8pt]
            \texttt{[One-Shot Example]}\\[8pt]
            \textbf{Llama Output:} \\
            
        }
    }
    \caption{\textbf{Prompt for Llama's answer generation and its results.}}
    \label{fig:llama_answer}
\end{figure}

The prompt for LLaMA-3.1-8B Instruct answer generation is shown in \cref{fig:llama_answer}. 
The scene description extracted from LLaVA, along with the corresponding prompt, is provided as input to Llama with an one-shot example.

\section{Llama Prompt for Language Evaluation}
\label{sec:llama_evaluation}
\begin{figure}[t]
    \centering
    \setlength{\fboxsep}{8pt}
    \fbox{
        \parbox{0.4\textwidth}{ 
            \small
            \textbf{Prompt:} \\
            \texttt{You will be given a sentence, and your task is to evaluate its fluency. Fluency refers to how natural and grammatically correct the sentence sounds in English. Rate the sentence on a scale from 1 to 10, where:} \\
            \texttt{1: Very unnatural, severe grammar or structural errors.} \\
            \texttt{10: Perfectly natural, indistinguishable from a native speaker's writing.} \\[8pt]
            \texttt{Your output should be in JSON format, containing your evaluation results based on the criteria above.}
        }
    }
    \caption{\textbf{Prompt for Llama evaluation of fluency scores for question and answer(QF and AF).}}
    \label{fig:llama_evaluation}
\end{figure}

The prompt for LLaMA evaluation is shown in \cref{fig:llama_evaluation}. This prompt is used to evaluate fluency scores for both questions and answers (QF and AF).

\end{document}